\pdfoutput=1

\documentclass[11pt]{article}

\usepackage[]{acl}

\usepackage{times}
\usepackage{latexsym}

\usepackage[T1]{fontenc}

\usepackage[utf8]{inputenc}

\usepackage{microtype}

\usepackage{inconsolata}

\usepackage{graphicx}
\usepackage{multirow}
\usepackage{array, makecell}
\usepackage{booktabs} 
\usepackage{adjustbox}
\usepackage{colortbl}

\title{\textit{Lived Experience Not Found:} LLMs Struggle to Align with Experts on Addressing Adverse Drug Reactions from Psychiatric Medication Use}

\author{Mohit Chandra\textsuperscript{1}, Siddharth Sriraman\textsuperscript{1}, Gaurav Verma\textsuperscript{1}, Harneet Singh Khanuja\textsuperscript{1},\\\textbf{Jose Suarez Campayo\textsuperscript{2}}, \textbf{Zihang Li\textsuperscript{3}}, \textbf{Michael L. Birnbaum\textsuperscript{4}}, \textbf{Munmun De Choudhury\textsuperscript{1}} \\
  \textsuperscript{1}College of Computing, Georgia Institute of Technology\\
  \textsuperscript{2}Hospital General Universitario Gregorio Marañón\\
  \textsuperscript{3}Hofstra University, \textsuperscript{4}Columbia University\\
  \small{
 \texttt{\{mchandra9, sidsr, gverma, hkhanuja3\}@gatech.edu; jsuarezc@salud.madrid.org}}\\
 \small{\texttt{zli56@pride.hofstra.edu; michael.birnbaum@nyspi.columbia.edu; munmun.choudhury@cc.gatech.edu}}}

\begin{document}

\definecolor{asparagus}{rgb}{0.53, 0.66, 0.42}
\definecolor{amaranth}{rgb}{0.9, 0.17, 0.31}

\def\mohit #1{\textcolor{red}{mohit: #1}}
\def\mdc #1{\textcolor{blue}{mdc: #1}}
\def\gv #1{\textcolor{asparagus}{gv: #1}}
\def\ss #1{\textcolor{orange}{ss: #1}}
\def\hsk #1{\textcolor{yellow}{hsk: #1}}

\def\dataset {Psych-ADR}
\def\framework {ADRA}

\maketitle
\begin{abstract}
Adverse Drug Reactions (ADRs) from psychiatric medications are the leading cause of hospitalizations among mental health patients. With healthcare systems and online communities facing limitations in resolving ADR-related issues, Large Language Models (LLMs) have the potential to fill this gap. Despite the increasing capabilities of LLMs, past research has not explored their capabilities in detecting ADRs related to psychiatric medications or in providing effective harm reduction strategies. To address this, we introduce the \textbf{\dataset} benchmark and the \textbf{A}dverse \textbf{D}rug Reaction \textbf{R}esponse \textbf{A}ssessment~(\textbf{\framework})~framework to systematically evaluate LLM performance in detecting ADR expressions and delivering expert-aligned mitigation strategies. Our analyses show that LLMs struggle with understanding the nuances of ADRs and differentiating between types of ADRs. While LLMs align with experts in terms of expressed emotions and tone of the text, their responses are more complex, harder to read, and only 70.86\% aligned with expert strategies. Furthermore, they provide less actionable advice by a margin of 12.32\% on average. Our work provides a comprehensive benchmark and evaluation framework for assessing LLMs in strategy-driven tasks within high-risk domains.
\end{abstract}

\section{Introduction}

Adverse Drug Reactions (ADRs)\footnote{An Adverse Drug Reaction (ADR) is defined as \textit{an appreciably harmful or unpleasant reaction, resulting from an intervention related to the use of a medicinal product, which predicts hazard from future administration and warrants prevention or specific treatment, or alteration of the dosage regimen, or withdrawal of the product.}~\cite{edwards2000adverse}.} caused due to psychiatric medications are a leading cause of hospitalizations among individuals with mental health conditions, accounting for 51.9\% to 91.8\% of cases, as reported in previous studies~\cite{angadi2020prevalence, ejeta2021adverse}. With nearly 70\% of
 
 \begin{figure}[!h]
    \centering
    \includegraphics[width=\columnwidth]{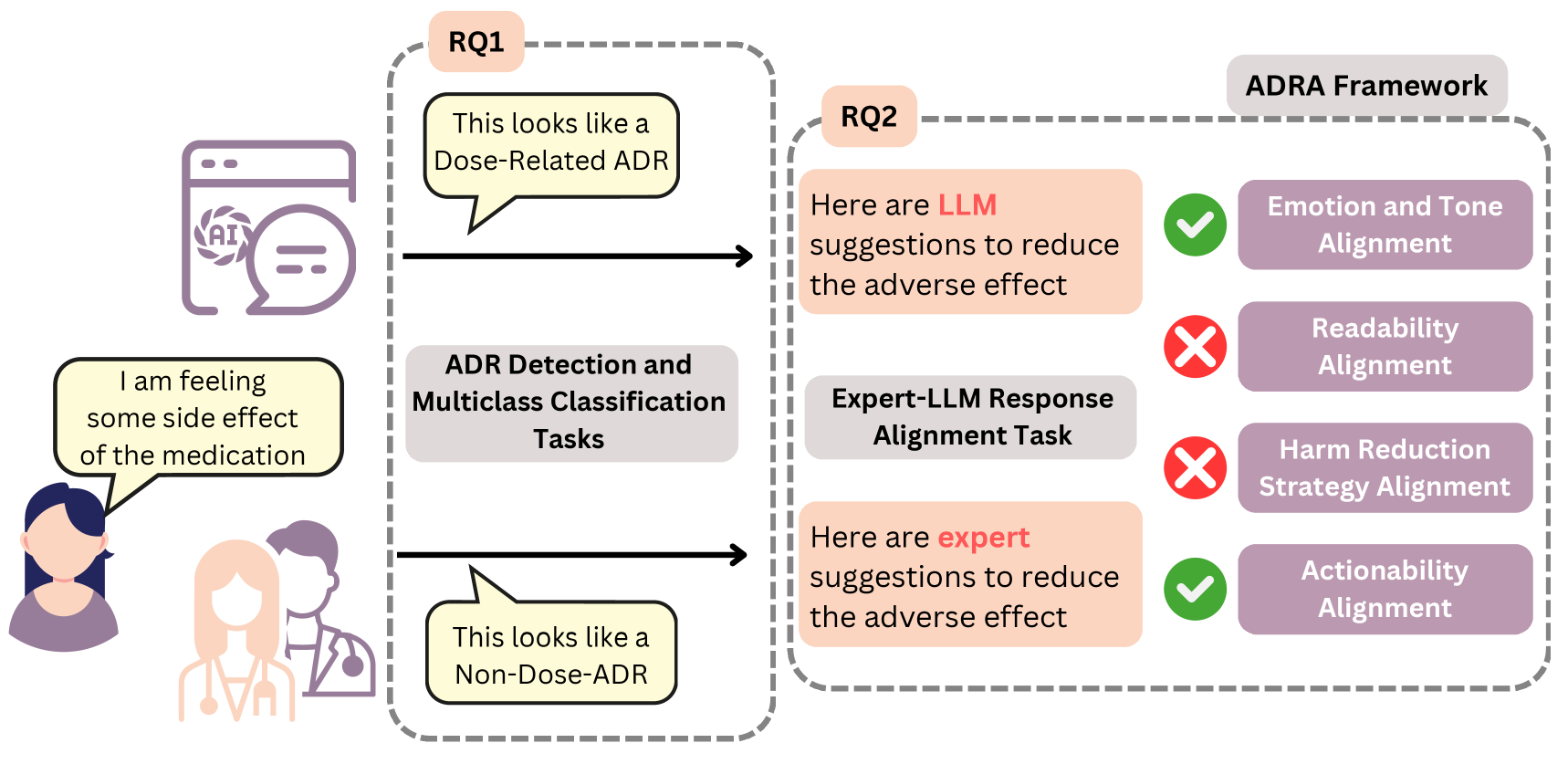}
    \caption{Overview of work; we present two tasks in this work -- ADR detection and multiclass classification (\textbf{RQ1}), and Expert-LLM response alignment (\textbf{RQ2}).}
    \label{fig:overview_figure}
    \vspace{-2em}
\end{figure}
 
\noindent individuals worldwide having limited or no access to mental health professionals~\cite{kazdin2013novel}, many patients increasingly turn to social media platforms such as Reddit to share experiences and seek advice~\cite{lee2017adverse, de2014seeking}. Yet, around 35\% of posts on mental health-related subreddits go unanswered, leaving many without adequate support~\cite{guimaraes2021comparing}. Further, while social media offers a platform for seeking assistance with resolving ADR queries, responses are frequently provided by individuals lacking expertise, raising concerns about the reliability of the information shared~\cite{vosoughi2018spread,wang2019systematic}. Hence, as conversational AI platforms (such as ChatGPT) gain prominence, more individuals are turning to these systems for healthcare-related queries, including those about psychiatric medication and ADRs.

Given the current limitations of healthcare systems and social media platforms, alongside the growing capabilities of LLMs in mental health-related tasks~\cite{yang-etal-2023-towards, yang2024mentallama, singhal2023towards}, LLMs have the potential to bridge the gap in online discussions by providing high-quality, contextual responses to ADR queries related to psychiatric medications. While previous studies have focused on detecting ADRs using deep learning methods, these efforts have primarily addressed non-mental health-related ADRs~\cite{mesbah2019training, sarker2015portable, karimi2015text}. Detecting ADRs related to psychiatric medications and evaluation of feedback responses from LLMs towards ADR-related queries, remains unexplored. This gap is particularly significant because addressing ADRs caused by psychiatric medications presents unique challenges, given the complex interplay between mental health conditions, their symptoms, and the potential for psychiatric medications to either alleviate or exacerbate those issues. Furthermore, for LLMs to meaningfully contribute to online discussions and effectively address ADR-related queries, it is essential to rigorously evaluate the quality of their long-form text responses and their alignment with expert knowledge in specialized domains such as psychiatry. This evaluation must consider the LLMs' ability to grasp the complexities of psychiatric medication ADRs and deliver responses that are contextually nuanced. Additionally, there is a need to evaluate the ability of LLMs to accurately portray the lived experiences\footnote{\textit{Personal knowledge about the world gained through direct, first-hand involvement in everyday events rather than through representations constructed by other people.}} of healthcare providers in addressing ADR-related queries which has been considered as an important factor in providing effective mental health support. In response to these needs and challenges, we address the following research questions:\vspace{0.5em}

\noindent\textbf{RQ1}: \textit{How effectively do LLMs detect concerns of ADRs associated with a broad range of psychiatric medications? Additionally, how accurate are LLMs in classifying different types of ADRs?}

\vspace{0.2em}

\noindent\textbf{RQ2}: \textit{To what extent do the responses from LLMs align with those from clinicians across different aspects when addressing ADR queries?}

\vspace{0.2em}

To answer the above stated research questions, we present \textbf{\dataset}~benchmark and \textbf{A}dverse \textbf{D}rug Reaction \textbf{R}esponse \textbf{A}ssessment~(\textbf{\framework})~framework. The proposed \dataset~benchmark includes 239 Reddit posts, labeled across two hierarchical levels for ADR detection and multiclass classification along with expert-written responses to queries. The proposed framework evaluates LLM-generated responses against those of medical experts, focusing on four assessment axes: (a) \textbf{text readability}, (b) \textbf{emotion and tone expression}, (c) \textbf{alignment of harm-reduction strategies}, and (d) \textbf{actionability of suggested strategies}.

\textbf{RQ1} results show that both ADR detection and ADR multiclass classification are challenging tasks, with the top model in a few-shot setting achieving F1 scores of 75.38 and 76.69 in respective tasks. We observed that all models exhibited a ``risk-averse'' nature, leading to a false-positive rate of over 70\%. Additionally, models struggled with non-dose-related and time-related ADRs, with GPT-4 Turbo misclassifying 51\% and 50\% of these instances, highlighting difficulties in grasping nuanced ADR types. For \textbf{RQ2}, LLM-generated responses were significantly harder to read than expert-written responses. In contrast, there was no observed significant difference in emotional or tone alignment between LLM and expert responses. However, the best model (OpenBioLLM-70B) achieved only 70.86\% alignment with expert harm-reduction strategies, and LLMs provided 12.32\% less actionable advice on average. Given the observations, our research has important real-world implications. The proposed benchmark provides a resource for evaluating LLMs on tasks involving the interaction between mental health conditions and psychiatric medications. The proposed framework hold practical utility for policymakers, practitioners, and healthcare professionals to assess LLM performance, especially in strategy-driven tasks in high-risk domains. The code repository for this work can be found here\footnote{\url{https://github.com/mohit3011/Lived-Experience-Not-Found}}.

\section{Data Collection and Curation}

We begin by providing the details of the data collection and filtering pipeline. We used publicly available data in English from Reddit spanning a one-year period (January 2019 - December 2019) obtained from Pushshift Reddit Dataset~\cite{baumgartner2020pushshift}. While the broad timeframe ensures large enough data before filtering, the specific period also \textit{(a)} predates the use of generative AI in day-to-day lives and \textit{(b)} the knowledge cutoff for all LLMs used in our evaluation. This allows for a fairer comparison between human experts and LLMs and also ensures minimal presence of machine-generated content on Reddit.

Following the past work~\cite{mesbah-etal-2019-training,saha2019social,chancellor2019discovering}, we selected 10 subreddits that focus on mental health-related issues or provide a platform for users to ask medical queries (such as r/depression and r/askdocs; see Appendix~\ref{appendix_sec:subrredit_list} for the complete list). To extract relevant posts, we compiled a set of 297 FDA-approved psychiatric medications provided by~\citet{saha2019social}. Further, to detect expressions of adverse symptoms in post titles and texts, we employed HealthE~\cite{gatto2023healthe}, a specialized named entity recognizer for identifying healthcare and medical entities. By combining the psychiatric medication names with the entities given by HealthE, we obtained 19,252 Reddit posts.

Filtering based on mentions of psychiatric medications and adverse symptoms provides a rich sample to extract posts that strictly discuss symptoms caused by psychiatric medications, which is the focus of our study. To specifically filter out posts expressing concerns of adverse drug reactions (ADRs), we prompted GPT-3.5 using definitions and specific conditions identified in previous research while also including insights from co-authors who are medical experts~\cite{edwards2000adverse}; see Appendix 
\ref{appendix_sec:adr_detection_prompts_criteria} and~\ref{appendix_sec:adr_multiclass_definitions} for complete list of criteria and exact prompts. Based on the annotations by GPT-3.5, we obtained 6,108 Reddit posts expressing ADR and 11,999
expressing no ADR (rest were deleted). Next, we discuss human validation of the labels for constructing the~\dataset~benchmark.

\section{The \dataset~Benchmark}
\noindent\textbf{LLM-assisted expert annotations}: We conducted expert-led annotations to validate the ADR labels generated by GPT-3.5 and to categorize the specific type of ADR described in each post, if applicable. Given the complexity and time-intensive nature of the human annotation process, we randomly selected 250 posts—consisting of both ADR-labeled and no-ADR-labeled posts as identified by GPT-3.5 for experts to annotate. Based on our discussions with the collaborating medical experts and drawing on the classification provided by~\citet{edwards2000adverse}, we categorized the ADRs into five granular types-- 1) dose-related ADR, 2) non-dose ADR, 3) dose- and time-related ADR, 4) time-related ADR, and 5) withdrawal ADR. We collaborated with three expert annotators --- two doctors, and one medical student, all with backgrounds in psychiatry with high proficiency in English. Based on the criteria provided for classifying a post as expressing ADR (Appendix~\ref{appendix_sec:adr_detection_prompts_criteria} and~\ref{appendix_sec:adr_multiclass_definitions}), they annotated each post to determine whether the post described an ADR, and if so, which category of ADR it belonged to along with providing reasoning for it (details related to the annotation tool in Appendix~\ref{appendix_sec:expert_annotation_task_details}).

The annotation task proved to be challenging for the annotators, with an average time of $\sim$7.2 minutes taken to annotate each post due to the complexity and subjectivity inherent in detecting adverse drug reactions. All three annotators agreed on the labels for 48\% of the posts. To address the disagreements, we conducted a second round of annotations in which all three annotators collaboratively resolved disagreements, resulting in the final set of labels (details in Appendix~\ref{appendix_sec:expert_annotation_task_details}). Finally, 11 posts were discarded due to their lack of relevance to ADRs, resulting in a final benchmark comprising 239 annotated posts. Table~\ref{tab:dataset_stats} presents the statistics for the~\dataset~benchmark.

\begin{table}[!h]
    \centering
    \begin{adjustbox}{width=\columnwidth}
    \begin{tabular}{l|r}
        \toprule
       \textbf{Class Label}  & \#\textbf{Examples}  \\
       \midrule
        \textbf{No-Adverse Drug Reaction} & 106 (44.4\%)\\
        \textbf{Adverse Drug Reaction (ADR)} & 133 (55.6\%)\\
        \hspace{5mm}Non-Dose ADR & 93 (38.6\%)\\
        \hspace{5mm}Withdrawal ADR & 22 (9.2\%)\\
        \hspace{5mm}Dose Related ADR & 13 (5.4\%)\\
        \hspace{5mm}Time Related ADR & 4 (1.7\%)\\
        \hspace{5mm}Dose and Time Related ADR & 1 (0.4\%)\\
        \bottomrule
    \end{tabular}
    \end{adjustbox}
\caption{Class-wise distribution of examples in the\dataset~benchmark dataset; 
\% w.r.t. $N = 239$.}
    \label{tab:dataset_stats}
        \vspace{-3mm}
\end{table}

\begin{table*}[h]
\centering
\begin{adjustbox}{width=\linewidth}
\begin{tabular}{l c c c c c c c c c c c c}
\toprule
& \multicolumn{6}{c}{\textbf{ADR Detection}} & \multicolumn{6}{c}{\textbf{ADR Multiclass Classification}} \\
\cmidrule(lr){2-7} \cmidrule(lr){8-13}
& \multicolumn{2}{c}{\textbf{Zero-Shot}} & \multicolumn{2}{c}{\textbf{5-Shot-Most Similar}} & \multicolumn{2}{c}{\textbf{5-Shot-Least Similar}} & \multicolumn{2}{c}{\textbf{Zero-Shot}} & \multicolumn{2}{c}{\textbf{5-Shot-Most Similar}} & \multicolumn{2}{c}{\textbf{5-Shot-Least Similar}} \\
\cmidrule(lr){2-3} \cmidrule(lr){4-5} \cmidrule(lr){6-7} \cmidrule(lr){8-9} \cmidrule(lr){10-11} \cmidrule(lr){12-13}
\textbf{Model} & \textbf{Acc.} & $\mathbf{F_1}$ & \textbf{Acc.} & $\mathbf{F_1}$ & \textbf{Acc.} & $\mathbf{F_1}$ & \textbf{Acc.} &$\mathbf{F_1}$ & \textbf{Acc.} & $\mathbf{F_1}$ & \textbf{Acc.} & $\mathbf{F_1}$ \\
\midrule
\rowcolor[HTML]{EFEFEF}
GPT-4 Turbo & \underline{72.03} & 68.55 & \underline{72.46} & \underline{69.52} & 72.46 & 70.05 & \textbf{57.58} & \textbf{62.16} & 65.91 & 69.36 & 60.61 & 64.42 \\
GPT-4o & \underline{72.03} & \underline{69.67} & 71.19 & 67.60 & 69.92 & 66.71 & 45.46 & 47.92 & 59.85 & 64.06 & 58.33 & 62.23 \\
\rowcolor[HTML]{EFEFEF}
Llama 3.1-70B Instruct & 69.88 & 65.34 & 71.97 & 69.11 & 71.97 & 69.29 & 48.87 & 52.55 & 64.66 & 69.15 & 62.41 & 66.88 \\
Llama 3.1-405B Instruct & 71.55 & 68.16 & 69.88 & 65.83 & \underline{73.22} & 70.91 & 46.62 & 50.31 & \textbf{74.44} & \textbf{76.69} & 65.41 & 69.15 \\
\rowcolor[HTML]{EFEFEF}
Claude 3 Haiku & 56.49 & 42.34 & 64.02 & 56.91 & 65.27 & 60.74 & 32.33 & 34.34 & 70.68 & \underline{76.41} & 54.14 & 62.00 \\
Claude 3 Opus & \textbf{77.41} & \textbf{76.44} & \textbf{76.57} & \textbf{75.38} & \textbf{75.73} & \textbf{74.39} & 42.11 & 44.68 & 69.93 & 73.79 & 63.16 & 68.28 \\
\rowcolor[HTML]{EFEFEF}
Claude 3.5 Sonnet & 68.62 & 63.48 & 71.13 & 68.18 & \textbf{75.73} & \underline{73.49} & 51.13 & 55.45 & 70.68 & 73.87 & 66.92 & \textbf{72.66} \\
OpenBioLLM-Llama3-70B & 61.09 & 50.92 & 70.71 & 67.43 & 69.04 & 66.17 & 37.59 & 37.08 & 71.43 & 73.88 & \underline{67.67} & 71.02 \\
\rowcolor[HTML]{EFEFEF}
Llama3-Med42v2-70B & 60.25 & 49.34 & 64.02 & 56.53 & 64.02 & 56.53 & \underline{56.40} & \underline{60.32} & \underline{72.18} & 75.16 & \textbf{68.42} & \underline{71.45} \\
\bottomrule
\end{tabular}
\end{adjustbox}
\caption{Performance of different models on Binary Detection and Multiclass Classification tasks under Zero-Shot and 5-Shot scenarios. We report the accuracy score (\textbf{Acc.}) and weighted $F_{1}$ score as ($\mathbf{F_1}$) with the best and second-best performing model metrics in each scenario highlighted in \textbf{bold} and \underline{underline}, respectively.}
\label{tab:adr_detection_performance}
\end{table*}

\noindent\textbf{Expert responses to ADR posts}: A key aspect of \dataset~benchmark is the inclusion of expert-written responses to queries in the ADR labeled posts. For each post that expressed an ADR related query, the most experienced annotator (Doctor) provided responses addressing the queries. To facilitate this, we identified and articulated the logical structure of the responses typically seen in clinical settings while working with the medical experts. In accordance to this structure, each response in our dataset begins with empathizing with the patient, followed by information on diagnosis, request for additional information, proposing harm reduction strategies to mitigate the ADR, and concluding with a final set of questions. An example response is shown in Figure~\ref{fig:appendix_dataset_answer_writing_sample} in the Appendix.

\section{Model Selection \& Implementation}

We conduct our analysis for the research questions with a total of 9 proprietary and open-weights LLMs. For proprietary models, we evaluate GPT-4o~\cite{gpt4o}, GPT-4 Turbo~\cite{achiam2023gpt}, Claude 3.5 Sonnet, Claude 3 Opus, and Claude 3 Haiku~\cite{anthropicIntroducingClaude}. For open-weights models, we evaluate LLama-3.1 405B Instruct-Turbo, LLama-3.1 70B Instruct-Turbo~\cite{llama31} and specialized medical LLMs -- Llama3-Med42-v2 70B~\cite{med42v2} and Llama3-OpenBioLLM 70B~\cite{openbiollm}. The choice of these models stems from their reported performance in different general-purpose and medical benchmarks~\cite{abbas2024comparing,nori2023can, chen2023meditron, anthropicIntroducingClaude}.

Previous studies have recommended lower temperatures for detection and labeling tasks to ensure more consistent outputs, while higher temperature values aid in more flexible generation~\cite{10.1145/3589334.3645643, achiam2023gpt}. Accordingly, for the ADR detection and multiclass classification tasks we set the temperature $t = 0$ and use $t=0.6$ for the response generation tasks. Beyond the task-specific temperature variations, the settings were kept consistent across all the LLMs. Additional details regarding the models, evaluation setup, and compute are provided in Appendix~\ref{appendix_sec:compute_details}.

\section{RQ1: Detecting Adverse Drug Reaction}
\label{sec:rq1_adr_detection}

For this task, we evaluated LLMs on detecting expressions of adverse drug reactions using the \dataset~benchmark. The evaluation involved two separate tasks: (1) identifying the presence or absence of concerns realted to ADRs in the 239 Reddit posts, and (2) classifying the type of ADR into one of five pre-defined categories for the 133 instances labeled as expressing ADRs in \dataset~benchmark. In both tasks we evaluated models using the zero-shot and few-shot variants of the chain-of-thought (CoT) prompting~\cite{wei2022chain}. Detailed prompts and classification criteria are provided in Appendix~\ref{appendix_sec:adr_detection_prompts_criteria} and~\ref{appendix_sec:adr_multiclass_definitions}.

Due to the wide variety of medications and symptoms in~\dataset~benchmark, we evaluated two different example sampling strategies for few-shot prompting. For this, we generated text embeddings for each Reddit title and post using Text-embedding-3-large~\cite{openaitextembedding}. Using cosine similarity, we retrieved the five most-similar and five least-similar posts for each example. Table~\ref{tab:adr_detection_performance} presents the accuracy and weighted $F_{1}$ scores for models in the ADR detection and ADR multiclass classification tasks.

\subsection{Zero-shot prompting on~\dataset}
\label{sec:rq1_results_zero_shot}

\noindent\textbf{Larger models typically perform better for ADR detection tasks, but this trend does not hold for ADR multiclass classification.} As expected, larger models (by parameter size) outperformed their smaller counterparts in the ADR detection task within their respective families, with Claude 3 Opus achieving the highest accuracy at 77.41\%, followed by GPT-4o and GPT-4 Turbo at 72.03\%. Interestingly, specialized medical models (OpenBioLLM-Llama3-70B and Llama3-Med42v2-70B) struggled in this task. However, for ADR multiclass classification, we did not observe any clear pattern between model size and performance. GPT-4 Turbo was the best performing model with an accuracy of 57.58\%, followed by Llama3-Med42v2-70B at 56.40\%. All models struggled with multiclass classification, likely due to the complexity of distinguishing between ADR types. Additionally, aligning with prior research, observed results in the multiclass classification showed that larger models do not always excel in specialized tasks~\cite{kanithi2024medic}.
\vspace{2pt}

\noindent\textbf{Models exhibited a ``risk-averse'' tendency, and prone to commit false-positive errors}. In both ADR detection and multiclass classification tasks, all models displayed ``risk-averse" behavior, often mislabeling posts without ADRs as positive for ADRs (see Appendix~\ref{appendix_sec:rq_1_adr_detection_error_analysis} for qualitative analysis). In zero-shot settings, Claude 3 Opus had a false-positive rate of 42\% for `ADR-No' labels, while Claude 3 Haiku's false positive rate was as high as 97\% (see Appendix~\ref{appendix_sec:rq_1_additional_results}). Similarly, in ADR multiclass classification, models struggled to distinguish between non-dose-related, dose-related, and time-related ADRs. GPT-4 Turbo misclassified 51\% of non-dose-related and 50\% of time-related ADRs in zero-shot settings. This risk-averse tendency indicates a lack of nuanced understanding of ADR complexities, which could lead to (a) patients discontinuing treatment~\cite{horne1999patients, horne2005concordance}, (b) increased fear about their conditions~\cite{starcevic2013cyberchondria}, and (c) ``alert-fatigue'' among healthcare providers~\cite{phansalkar2013drug}.

\subsection{Few-shot prompting on~\dataset}
\label{sec:rq1_results_few_shot}

\noindent\textbf{In-context learning enhances model performance but not in every case}. We observed the in-context learning in general improved performance of models for both ADR detection and multiclass classification tasks, with a more significant impact on the latter task. For multiclass classification, we observed an average increase of 18.14 and 23.06 points in weighted $F_1$ score among model performance using least-similar and most-similar example prompting respectively. However, this pattern was not observed in ADR detection task. Claude 3 Opus outperformed other models in the ADR detection, achieving an $F_1$ score of 76.44 with zero-shot prompting. In ADR multiclass classification, Llama-3.5-405B performed best with most-similar examples ($F_1$ 76.69). For analyzing the impact of providing examples in the ADR detection task, we observed that some models, such as Claude 3 Haiku showed an average improvement of $F_1$ score (16.49 points), whereas we did not observe such a trend for models such as GPT-4o, Claude 3 Opus in few-shot settings. The stochastic nature of LLM generation, coupled with the inability to learn nuances from examples in the "ADR-No" class, may be a contributing factor to this issue. This was further confirmed as we noted that even in few-shot settings, models exhibited "risk-averse" behavior with high false-positive rates, indicating that providing examples could not effectively compensate for the lack of ``lived-experience'' in the models. This was the major reason behind models failing to achieve the expected gains in detecting ADR.

\noindent\textbf{Impact of choosing similar or diverse examples depends upon the task}. While the performance boost in the ADR multiclass classification task could be attributed to the predominance of non-dose-related ADRs, the comparatively smaller performance gains observed when models were presented with the five least similar examples suggest that models were able to grasp the contextual information presented through the examples and capture the nuances of various ADR types. However, no such pattern was observed in the ADR detection task, with 3 models showing increase in $F_1\geq1\%$ with five least similar examples based prompting. This showed that diversity in examples rather than stochasticity impacted model performance.

\section{RQ2: Alignment between human and AI feedback}
\label{sec:rq2_human_ai_alignment}

Evaluation of long-form text generation is an open problem and involves many challenges like isolating the stylistics from the semantics. However, in the context of responses to ADR queries, we propose abstracting out the LLM generations and ground-truth expert responses to four key components -- (1) emotion and tone, (2) text readability, (3) harm reduction strategy, and (4) actionability of proposed strategies. Via this abstraction to key components, our alignment evaluations focus on specific aspects that contribute towards an ideal response to ADR queries. We explain the importance of these components below and the methodology for evaluation.
\vspace{2pt}

\noindent\textbf{Emotional and tone alignment}: Emotional intelligence is regarded as a key factor in healthcare, fostering strong therapeutic relationships that drive meaningful change~\cite{king2011structure}. Therefore, LLM-generated responses should align with expert-written responses in terms of tone and expressed emotion. To assess this, we used Empath~\cite{fast2016empath}, a widely-used lexicon-based tool, focusing specifically on 8 relevant emotional and tonal categories identified from prior literature~\cite{riess2014empathy, mechanic2000concepts}. We analyzed the distribution of these categories in LLM-generated and expert responses, and quantified their differences using Kullback-Leibler (KL) divergence to measure alignment of expressed emotions and tone in the LLM and expert responses.
\vspace{2pt}

\noindent\textbf{Text readability alignment}: Past studies have shown that health literacy is strongly correlated with patient outcomes~\cite{wolf2005health}. A major factor contributing to lower health literacy is the communication barrier between patients and healthcare providers, which often arises from the complexity of medical text, including the writing style and choice of terminology~\cite{dubay2004principles}. Hence, the responses produced by LLMs should be easily readable and be of comparable to that of the expert-written responses. To assess this, we used SMOG index~\cite{mc1969smog}, a popular readability index to assess health literacy material.
\vspace{2pt}

\noindent\textbf{Harm reduction strategy alignment}: In cases of adverse reactions to psychiatric medications, suggesting safe medical interventions is crucial to prevent further harm. We operationalized these interventions using \textit{harm reduction strategies} (HRS) ~\cite{hrs1995}, aimed at minimizing the negative effects of medications that one is reliant on. Ideally, LLMs should propose strategies that align with the expert's responses.

To compare the harm reduction strategies suggested by LLMs and experts, we took inspiration from methods for entailment and factuality evaluation in long-form texts~\cite{factscore, deepmind-safe, kamoi-etal-2023-wice}. First, we extracted atomic HRS from LLM responses by prompting GPT-4o~\cite{gpt4o}. Since some extracted strategies were redundant, we used a few-shot approach to combine those that suggested the same overall approach but differed in specific details to get the final set of HRS for each response (examples in Table~\ref{tab:hrs_combination_examples}). To check for the robustness of the extraction and combination method, we conducted a round of human evaluation with 4 annotators. Using a random sample of 40 responses for each task, we evaluated 193 strategies for the extraction and 174 strategies for combination, and obtained a correlation score of 92\% and 90\% respectively with LLM evaluation.

We then evaluated alignment of HRS for each LLM-expert response pair using two methods. First, we used AlignScore \cite{zha-etal-2023-alignscore}, a widely-used text alignment method providing a score between 0 and 1 based using a fine-tuned RoBERTa-large model \cite{liu2019robertarobustlyoptimizedbert}. We computed AlignScore for each strategy from the LLM response against the expert response. We obtained a response-level AlignScore by averaging the scores across all HRS for the response. Second, for a more interpretable alignment score, we prompted GPT-4o with in-context examples to reason and classify if a strategy is aligned with the expert's response. We computed a response-level GPT-4o score by computing the percentage of aligned HRS over total number of HRS. These two approaches ensured robustness by covering both a continuous alignment score and one based on reasoned binary alignment labels. We conducted another round of human evaluation for the GPT-4o score, where four annotators annotated 40 responses, achieving a 95\% correlation with GPT-4o's score and reasoning. Prompts for LLM-based tasks are presented in Table \ref{tab:appendix_hrs_extraction_prompt}, \ref{tab:appendix_hrs_combination_prompt} \& \ref{tab:appendix_hrs_alignment_prompt}, and human evaluation details are presented in Appendix \ref{appendix_sec:hrs_annotation_correlation}.
\vspace{1pt}

\noindent\textbf{Actionability alignment}: Prior work in health communication has recognized the importance of \textit{actionability} in the responses of healthcare professionals to enable greater engagement and encourage increased action from patients~\cite{sharma2023cognitive}. To this end, we designed an approach to measure the alignment between LLM responses and expert responses along the actionability dimension. We first decomposed actionability into specific sub-dimensions while working with clinical experts and using the guidelines presented in the Patient Education Materials Assessment Tool (PEMAT;~\citeauthor{pematguidelines}). Harm reduction strategies recommended by experts and LLMs should be: (i) practical, (ii) contextually relevant, (iii) specific, and (iv) clear. We present concrete definitions for each of the sub-dimensions in Appendix~\ref{sec:appendix_actionability_criteria}.

To operationalize the quantification of actionability alignment, we prompted the GPT-4o model using carefully selected in-context learning examples and chain-of-thought prompting. The GPT-4o model considers the ADR post made by the user and assigns a binary label to each harm reductions strategy based on whether or not the target sub-dimension of actionability is present in the strategy (0: absent; 1: present). To validate the labels assigned by the GPT-4o model, the medical experts reviewed the rationales generated for detecting each of the sub-dimensions of actionability in $100$ harm reduction strategies, and agreed with 91 of them for practicality, 94 for relevance, 82 and 89 for specificity and clarity, respectively. Overall, the extent of the agreement between experts and GPT-4o rationales reinforced the validity of the labels assigned to the 4 sub-dimensions of actionability.  
Following this, for responses generated by the LLMs, we computed the fraction of harm reduction strategies that are aligned with the HRS \textit{and} also demonstrate presence of a certain sub-dimension of actionability. For instance, for the practicality dimension, the LLM-generated HRS are scored as: 

{\small
\vspace{-3mm}
\[
    \mathrm{Practicality}_{\mathrm{LLM}} = \frac{\mathrm{\#\ aligned\ \&\ practical\ HRS}}{\mathrm{\#\ total\ HRS}}
\vspace{-1mm}
\]}
It is worth emphasizing that the constraint of only considering aligned HRS within the LLM-generated responses enforces a penalty for generating unaligned HRS while computing actionability. Since expert responses are inherently always aligned, their HRS do not undergo such a penalization. We present the average scores for the 4 sub-dimensions and their aggregate as the overall actionability score in Table \ref{tab:actionability_scores}.

\subsection{Results}

\begin{figure}[!h]
    \centering
    \includegraphics[width=\columnwidth]{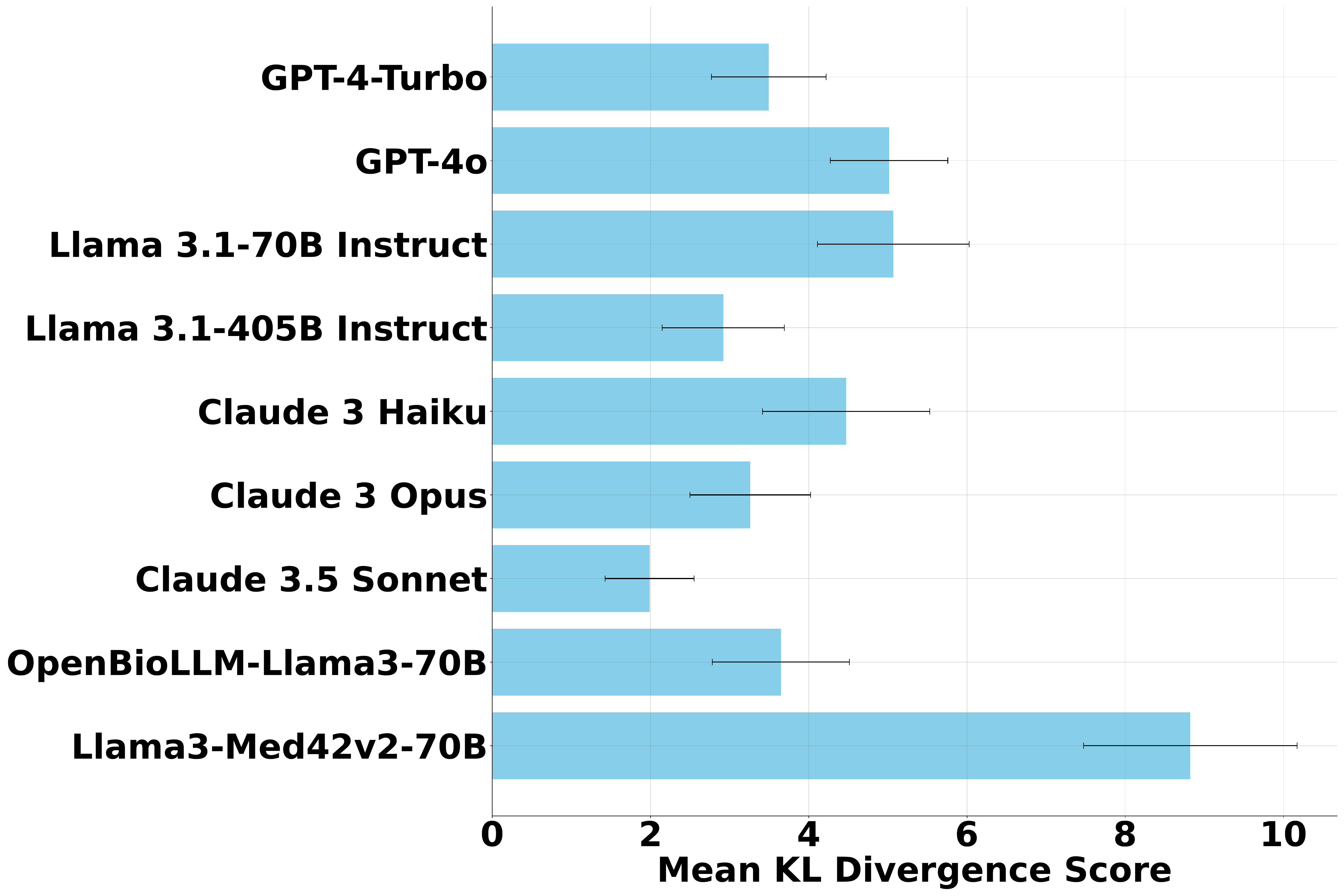}
    \caption{Mean KL Divergence score for the empath categories distribution between models and the expert responses in the \dataset~benchmark dataset. (Lower score is better).}
    \label{fig:empathy_mean_kl_divergence_score}
    \vspace{-3mm}
\end{figure}

\noindent\textbf{Emotional and tone alignment}. Figure~\ref{fig:empathy_mean_kl_divergence_score} presents the mean KL-divergence score for the distribution of 8 Empath categories between LLM responses and expert-written response. A $\chi^2$ test was conducted to assess the differences in category distributions, and the $p$-values were non-significant across all models, indicating that the models' responses were \textit{not} significantly different from the expert-written responses in terms of emotions expressed and the tone used. Further, we observed that larger and more capable models from the Llama and Claude families showed greater alignment with expert responses across different emotional and tone related categories. Interestingly, Llama-3 Med42v2 70B performed the worst. This could be attributed to the fact that a major portion of dataset used for instruction fine-tuning for this model was obtained from the medical and biomedical literature, which may not prioritize emotional communication while providing responses~\cite{med42v2}. 

Upon closer examination of the individual categories (Figure~\ref{fig:appendix_empath_category_distribution}), we found that the expert responses on average showed higher levels of anticipation and affection in the category distribution compared to LLMs. Similarly, a helping tone was more prominent in the expert responses in 6 out of 9 comparisons. However, LLMs exhibited higher use of optimistic and cheerful tones in their responses on average. Additionally, 6 out of 9 LLMs produced responses that used a more polite tone, incorporating more trust-based emotions.

\begin{figure}[!h]
    \centering
    \includegraphics[width=\columnwidth]{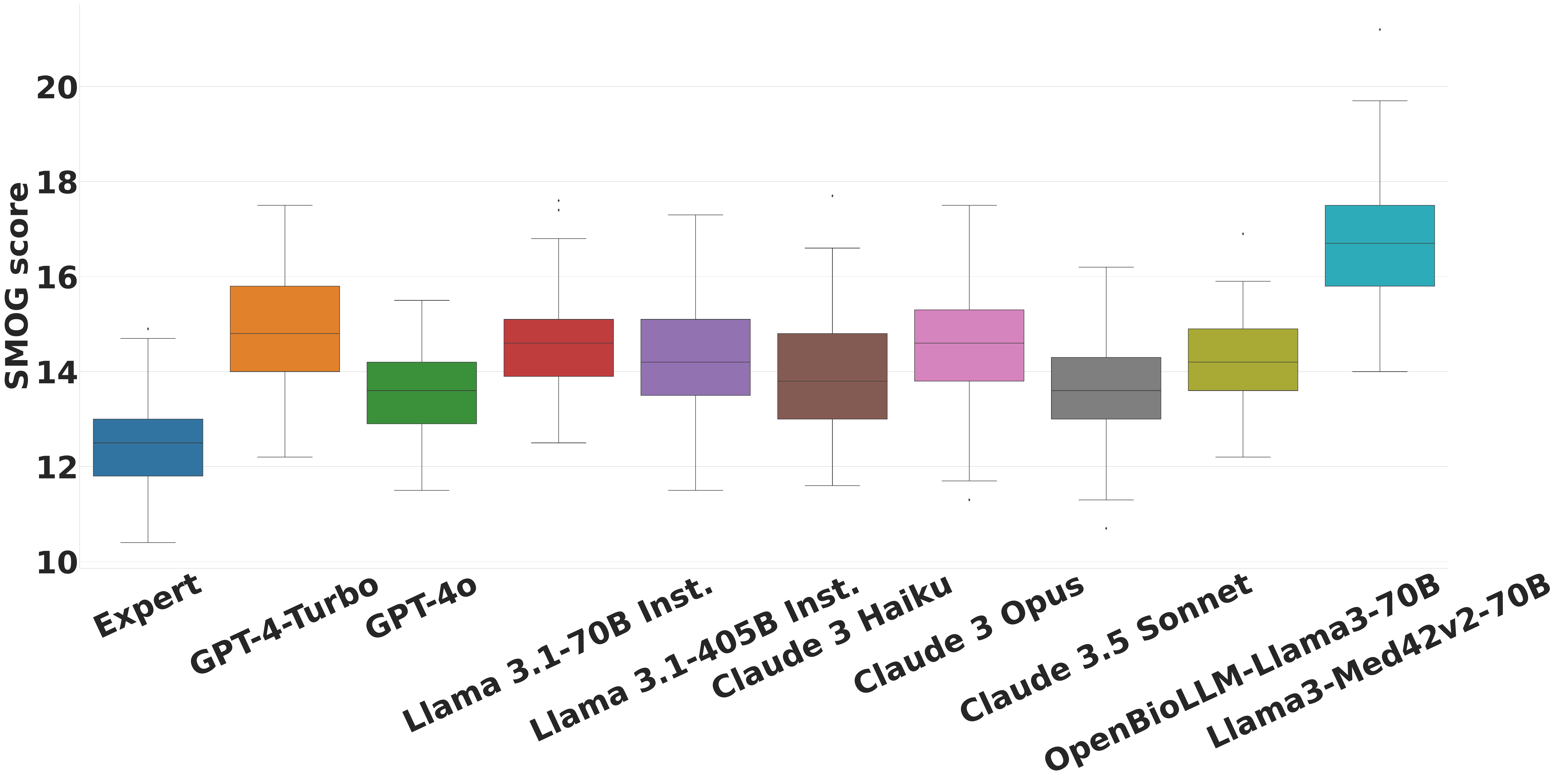}
    \caption{Mean SMOG Scores and 95\% Confidence Intervals for Various Models (lower values are better).}
    \label{fig:readability_results}
    \vspace{-10pt}
\end{figure}

\noindent\textbf{Text readability alignment}. Figure~\ref{fig:readability_results} presents the box plot for LLM and expert response SMOG scores. As observed, LLM-generated responses tend to be more complex, reflected in higher SMOG scores compared to those written by experts (SMOG$_{mean}$11.02) in the \dataset~benchmark. Welch's $t$-test~\cite{welch1947generalization} further revealed that the SMOG scores of expert-written responses were significantly lower than those of any LLM-generated responses. We also observed that more capable models produced more readable responses (with Claude 3 Opus being an exception). Similar to the findings on emotional alignment, Llama3-Med42v2-70B showed the lowest alignment with the expert-written responses, producing the most complex responses, likely due to a major portion of instruction-tuning data coming from medical and biomedical scientific literature. In contrast, OpenBioLLM-Llama3-70B outperformed many proprietary models, likely due to the custom dataset used for fine-tuning.

\noindent\textbf{Harm reduction strategy alignment}. Table~\ref{tab:hrs_alignment_performance} presents the mean response-level AlignScores and GPT-4o scores for alignment of harm reduction strategies of LLMs with the expert's responses. In line with results from zero-shot ADR classification in RQ1, more capable models Llama 3.1-405B Instruct and Claude 3.5 Sonnet in their respective families tended to produce strategies less aligned with the expert than their smaller counterparts, validating a previously observed pattern of LLM performance in responding to open-ended clinical questions \cite{kanithi2024medic}. While the open-weights models performed on par or better than proprietary models across both alignment metrics, the best-performing medical model (OpenBioLLM-Llama3-70B) aligned with expert harm reduction strategies for 70.86\% of the cases, highlighting the need for further fine-tuning for specialized domains such as psychiatry. Qualitative analysis of non-aligned HRS revealed that most focused on general lifestyle advice, such as maintaining a healthy diet and sleep routine, rather than addressing actions related to the involved medication (details in Appendix \ref{appendix_sec:hrs_alignment_error_analysis}).

\begin{table}[!h]
\centering
\begin{adjustbox}{width=\columnwidth}
\begin{tabular}{l c c c c}
\toprule
& \multicolumn{2}{c}{\textbf{AlignScore}} & \multicolumn{2}{c}{\textbf{GPT-4o Score}} \\
\cmidrule(lr){2-3} \cmidrule(lr){4-5}
\textbf{Model} & \textbf{Mean} & \textbf{Std. dev.} & \textbf{Mean} & \textbf{Std. dev.} \\
\midrule
\rowcolor[HTML]{EFEFEF}
GPT-4 Turbo & 46.49 & 22.80 & \underline{65.28} & 27.22 \\
GPT-4o & 42.06 & 23.10 & 62.72 & 27.66 \\
\rowcolor[HTML]{EFEFEF}
Llama 3.1-70B Instruct & \underline{46.91} & 24.22 & 63.57 & 31.50 \\
Llama 3.1-405B Instruct & 39.96 & 20.70 & 54.71 & 32.30 \\
\rowcolor[HTML]{EFEFEF}
Claude 3 Haiku & 41.71 & 25.32 & 61.96 & 31.81 \\
Claude 3 Opus & 42.42 & 21.27 & 59.16 & 30.21 \\
\rowcolor[HTML]{EFEFEF}
Claude 3.5 Sonnet & 36.83 & 22.74 & 59.48 & 31.63 \\
OpenBioLLM-Llama3-70B & \textbf{56.55} & 24.81 & \textbf{70.86} & 30.46 \\
\rowcolor[HTML]{EFEFEF}
Llama3-Med42v2-70B & 42.59 & 22.47 & 61.61 & 28.19 \\
\bottomrule
\end{tabular}
\end{adjustbox}
\caption{Alignment of harm reduction strategies of various models with the expert's response. We report the mean and standard deviation for the AlignScore metric GPT-4o score, with the \textbf{best} (bold) and \underline{second-best} (underline) performing model in each metric highlighted.}
\label{tab:hrs_alignment_performance}
\end{table}

\noindent\textbf{Actionability alignment.} In Table \ref{tab:actionability_scores}, we noted that expert responses scored the highest on overall actionability in comparison to all the LLMs (0.46). Nonetheless, medical models like OpenBioLLM-Llama3-70B and Llama3-Med42v2-70B demonstrate reasonable actionability scores (0.44), followed by other proprietary and open-weights models (0.35 to 0.43). Beyond the aggregate actionability score, the scores for the sub-dimensions provide interesting insights on alignment between expert and LLM responses. While expert responses were rated considerably better than all LLM responses in terms of the practicality (0.83) and contextual relevance (0.73) of the harm reduction strategies, their specificity (0.17) and clarity (0.13) are relatively lacking. This indicates that while LLMs tend to demonstrate greater specificity and clarity in their harm reduction strategy, the recommended strategies may often not be feasible  and contextually relevant, considering the users' personal circumstances, such as physical ability, financial resources, and time constraints. This observation further reinforces the need of encoding and reflecting on lived experiences~\cite{de2023benefits, lawrence2024opportunities} as part of ADR responses to address contextual cues, a dimension along which LLMs need to improve further. 

\begin{table}[!h]
\centering
\begin{adjustbox}{width=\columnwidth}
\begin{tabular}[width=\columnwidth]{lccccc}
\toprule
\textbf{Model} & \textbf{Practical} & \textbf{Relevant} & \textbf{Specific} & \textbf{Clear} & \textbf{Actionable} \\
\midrule
\rowcolor[HTML]{CDC1FF}
Expert Responses & 0.83 & 0.73 & 0.17 & 0.13 & 0.46 \\
\rowcolor[HTML]{EFEFEF}
OpenBioLLM-Llama3-70B & 0.68 & 0.70 & 0.17 & 0.22 & 0.44 \\
Llama3-Med42v2-70B & 0.60 & 0.61 & 0.26 & 0.29 & 0.44 \\
\rowcolor[HTML]{EFEFEF}
Claude 3 Haiku & 0.64 & 0.64 & 0.21 & 0.24 & 0.43 \\
Claude 3.5 Sonnet & 0.63 & 0.61 & 0.20 & 0.22 & 0.42 \\
\rowcolor[HTML]{EFEFEF}
GPT-4 Turbo & 0.63 & 0.62 & 0.17 & 0.21 & 0.41 \\
Llama 3.1-70B Instruct & 0.62 & 0.64 & 0.17 & 0.18 & 0.40 \\
\rowcolor[HTML]{EFEFEF}
GPT-4o & 0.59 & 0.57 & 0.15 & 0.19 & 0.38 \\
Claude 3 Opus & 0.57 & 0.54 & 0.16 & 0.17 & 0.36 \\
\rowcolor[HTML]{EFEFEF}
Llama 3.1-405B Instruct & 0.58 & 0.56 & 0.13 & 0.14 & 0.35 \\
\bottomrule
\end{tabular}
\end{adjustbox}
\caption{Mean actionability alignment scores of HRS (last column), computed as average of practicality, relevance, specificity, and clarity scores.}
\label{tab:actionability_scores}
\vspace{-5mm}
\end{table}

\section{Related Work}
\label{sec:related_work}

\noindent\textbf{Large language models in healthcare}: With the growing capabilities of LLMs, past studies have explored their potential to assist stakeholders in healthcare domain. Proprietary models like GPT-4 and MedPalm have shown strong performance on multiple-choice benchmarks and even passed exams such as the USMLE~\cite{singhal2023large, singhal2023towards, nori2023capabilities, openbiollm, kanithi2024medic}. LLMs have also been evaluated for mental health support queries~\cite{yang-etal-2023-towards}. However, previous research has also highlighted challenges for LLMs in these settings, highlighting cross-lingual disparities~\cite{10.1145/3589334.3645643}, gender and geographic biases~\cite{restrepo2024analyzing}, and limitations in clinical competency tests for both general and mental health~\cite{thirunavukarasu2023trialling, jin2023psyeval}.

\noindent\textbf{ADR detection and pharmacovigilance}: Past research has looked into ADR detection through social media platforms~\cite{mesbah2019training,sarker2015portable,karimi2015text}. However, these studies have predominantly focused on non-mental health related cases, relying on binary classification tasks with limited medication datasets. In contrast, medical studies on ADRs related to psychiatric medications~\cite{angadi2020prevalence, ejeta2021adverse} are typically hospital-based, small-scale, and not focused on detecting ADRs within online communities.

\noindent\textbf{Importance of lived experience}: The importance of lived experience has been studied in various works across the field of mental health research, psychology, and education. Understanding lived experiences provides insight into individuals' personal realities and preferences, contributing towards a deeper understanding of their experiences, expectations and requirements. In mental health research, previous studies have highlighted the importance of understanding the experiences and realities of individuals living with mental health conditions for providing better treatment~\cite{gilbert2012wounded, repper2011review}.~\citet{byrne2013things} further highlighted that students showed positive attitudes and increased self-awareness towards the impact of mental illness on individuals when they were taught by people with lived experience of mental health challenges. Past research in psychology has also stressed on the understanding of the lived experience of individuals belonging to different backgrounds. Previous studies have also highlighted the importance of lived experience in the form of experiential knowledge among the healthcare provider for making decisions~\cite{lyu2023lived, palukka2021outlining}. In summary, previous research in mental health and psychology has emphasized on the multifaceted importance of lived experience among the patients, educators and healthcare providers.

\section{Conclusion and Future Work}

In this work, we proposed the~\dataset~benchmark and~\framework~framework for evaluating the alignment of LLMs with experts on responding ADR queries caused due to psychiatric medication use. In our  RQ1 analysis, even the best models achieved only 77.41\% accuracy in detecting ADR and 74.44\% accuracy in detecting the type of ADR. Our RQ2 analysis further revealed that while models align with experts on expressed emotions and tone of the text, they struggle in important areas like readability, alignment of harm reduction strategies with expert knowledge, and suggesting actionable interventions. Our work can inspire future work to adopt a more holistic approach for evaluating models, emphasizing the integration of ``lived experience" alongside expert knowledge.

\section{Broader Implications}

Responding to ADR queries is challenging due to the complexity of mental health conditions, symptoms, and medication effects.  The results from the analyses of RQ1 and RQ2 surface these challenges, revealing nuanced patterns that highlight the intricacies involved, and hence findings from this work have several key implications:

\noindent\textbf{Going beyond the choice-based medical benchmarks}. LLMs have achieved near-perfect scores on popular medical benchmarks~\cite{nori2023capabilities,singhal2023towards}, however, these evaluations typically focus on multiple-choice or case-based questions,which don't reflect the nuanced understanding required in real-world scenarios like mental health. Despite their strong performance on medical tasks, Llama3-Med42v2-70B and OpenBioLLM-Llama3-70B struggled with detecting ADRs and providing aligned and actionable HRS, highlighting the need to move beyond standard benchmarks towards more holistic alignment evaluation paradigms.\\
\noindent\textbf{Focusing on empowering experts rather than replacing them}. While LLMs did not match expert performance in our analysis, they showed a potential to enhance healthcare by providing clearer, more actionable responses. Given the global shortage of mental health professionals~\cite{kazdin2013novel}, LLMs could expand access to mental healthcare and support experts with further fine-tuning and alignment with expert reasoning.\\
\noindent\textbf{Disentangling inclusion of humanistic features in LLMs and advocacy for inclusion of lived experience}. While our work provides evidence of the lack of lived experience, which is essential for understanding the nuances of a complex task such as ADR detection and for proposing mitigation strategies in model responses, we do not advocate for increasing human-like features in LLMs. Previous studies have suggested that heightened anthropomorphism, independent of whether it is accompanied by enhanced capabilities, can increase trust among individuals~\cite{natarajan2020effects, chen2021anthropomorphism}. Hence, developers and researchers need to be cautious before introducing such features as individuals may trust LLM responses even when they provide incorrect or inconsistent information which can be hazardous in high-risk domains such as healthcare. In contrast, we advocate for approaches that align with previous research, which has shown that the efficacy of LLMs in the healthcare domain can be enhanced through fine-tuning on specialized data or by incorporating useful features into the model, without introducing human-like features~\cite{belyaeva2023multimodal, li2023chatdoctor}.

\section{Limitations}

While novel, it is important to acknowledge the limitations of our work. While the proposed \dataset~benchmark is the first to focus exclusively on ADRs related to psychiatric medications, the number of examples used for evaluating LLMs is limited, which may not capture the full range of ADRs associated with these medications. We also acknowledge that expanding a benchmark like ours presents challenges due to the time-intensive nature of annotation and response writing, compounded by the subjectivity and complexity inherent in this domain. Additionally, while the responses were provided by a highly experienced doctor, variations in clinical opinions are possible given the subjective nature of ADR assessment in psychiatric medication contexts. Despite these limitations, we believe that the proposed~\dataset~benchmark provides a valuable resource for further research, offering a robust starting point for the study of ADRs in psychiatric medications.

We also acknowledge certain limitations in the~\framework~framework. Although we aimed to compare responses across a set of relevant emotions and tones, our approach relies on a lexicon-based method, which may sometimes miss semantic meaning of the responses. Additionally, the harm reduction strategy alignment in our framework excludes strategies suggested by LLMs that are not present in the expert responses. However, there may be cases where the LLM's proposed strategy is a viable option according to other clinicians, but due to the open-domain nature of the problem and the lack of a verified data source, we were unable to evaluate the correctness of such strategies. Despite these challenges, our work provides a robust framework for assessing the capabilities of LLMs in high-risk strategy-driven domains.

\section{Ethical Considerations}
\label{sec:ethical_considerations}

We collected public domain social media data from a publicly available dataset which allowed us to use the resource for non-commercial purposes. We further ensured that all data used was de-identified and did not contain any offensive content. As our study involved working with retrospective data without direct interaction with the authors of the posts, the Institutional Review Board (IRB) classified it as non-human subjects research, exempting it from IRB approval. Still, we adhered to established best practices for working with social media data, as recommended in the literature~\cite{weller2016manifesto, weller2015uncovering}. In line with Reddit's data-sharing guidelines and relevant data-use agreements, we will provide access to the benchmark exclusively comprising Post IDs and annotations, to interested researchers upon acceptance. Further, we will make the code used in this work publicly available on a GitHub repository upon acceptance.

Our study presents a systematic approach for evaluating LLMs for addressing ADR related queries from psychiatric medication use, and hence does not inherently pose direct risks. However, it is important to emphasize that better performance on \dataset~benchmark should not be interpreted as an indication of increased capabilities in real-world applications. Instead, these results should be complemented with thorough human evaluation to ensure the reliability and safety of models.

\section{Acknowledgment}

Chandra and De Choudhury were partly supported through National Science Foundation (NSF) grant \#2230692, a grant from the American Foundation for Suicide Prevention (AFSP). Further, this research project has benefitted from the Microsoft Accelerate Foundation Models Research (AFMR) grant program. The findings, interpretations, and conclusions of this paper are those of the authors and do not represent the official views of NSF, AFSP or Microsoft. We further acknowledge the contributions from Dr. Santiago Alvarez Lesmes towards data annotation and assessment.

\bibliography{main}

\appendix
\clearpage

\begin{figure*}[t]
    \centering
    \includegraphics[width=\textwidth]{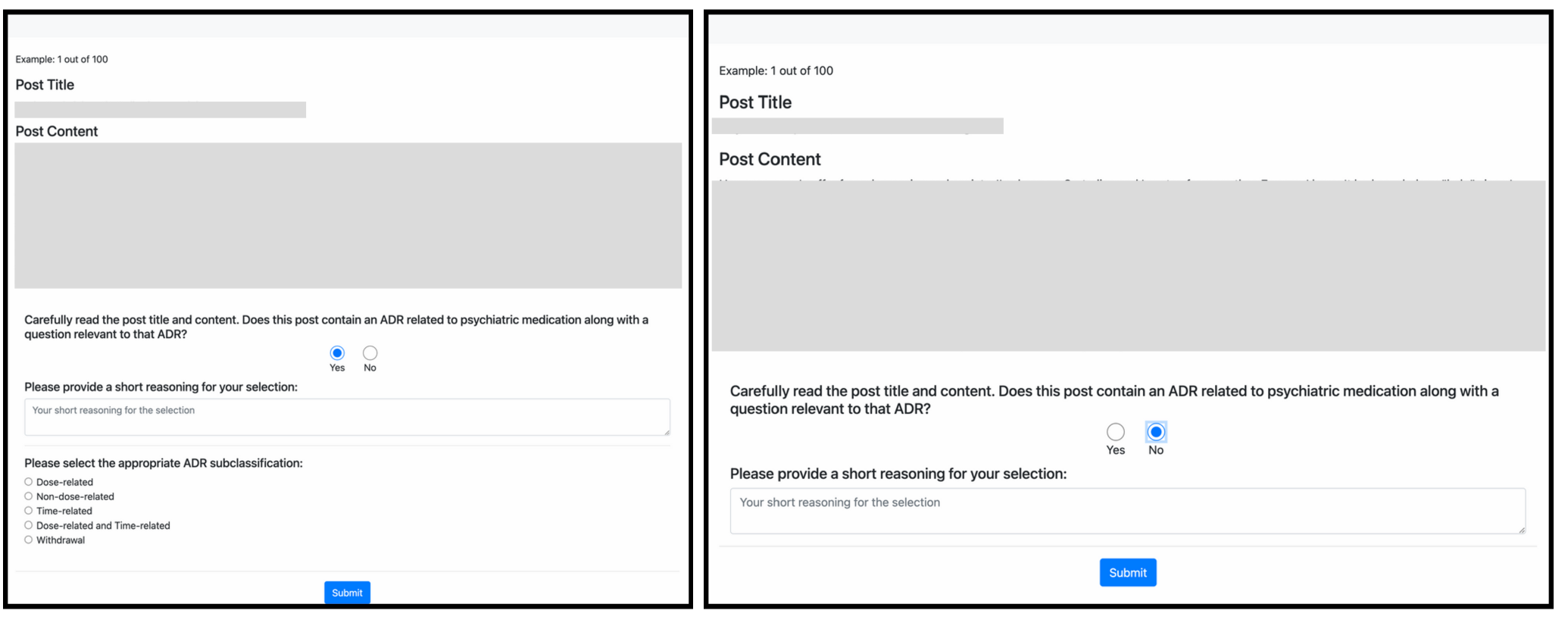}
    \caption{Annotation interface for the~\dataset~benchmark used in the annotation process. The interface displays the post title and content, along with access to annotation guidelines. In the left image screenshot, the annotator identifies an adverse drug reaction (ADR) related to psychiatric medication, then provides a brief rationale and selects the class of ADR. In the right image screenshot, the annotator indicates that no ADR is present, in which case only a rationale for this decision is required.}
    \label{fig:appendix_annotation_platform_screenshots}
\end{figure*}

\begin{table*}[t]
\centering
\small
\begin{tabular}{|l|l|l|}
\hline
\textbf{LLM} & \textbf{Version} & \textbf{Parameter Size} \\ \hline
GPT-4o & 2024-08-06~\cite{gpt4o} & (undisclosed) \\ \hline
GPT-4 Turbo & turbo-2024-04-09~\cite{achiam2023gpt} & (undisclosed) \\ \hline
Claude 3.5 Sonnet & 2024-06-20 & (undisclosed) \\ \hline
Claude 3 Opus & 2024-02-29 & (undisclosed) \\ \hline
Claude 3 Haiku & 2024-03-07~\cite{anthropicIntroducingClaude} & (undisclosed) \\ \hline
LLama-3.1 405B Instruct-Turbo & ~\cite{llama31} & 405 billion\\ \hline
LLama-3.1 70B Instruct-Turbo & ~\cite{llama31} & 70 billion\\ \hline
Llama3-Med42-v2 70B & ~\cite{med42v2} & 70 billion\\ \hline
Llama3-OpenBioLLM 70B & ~\cite{openbiollm} & 70 billion\\ \hline
\end{tabular}
\caption{Models and their corresponding version dates}
\label{tab:appendx_model_details}
\end{table*}

\section{List of subreddits}
\label{appendix_sec:subrredit_list}

We use the following list of subreddit to collect data for further filtering: `r/depression', `r/anxiety', `r/bipolar', `r/BPD', `r/schizophrenia', `r/autism', `r/mentalhealth', `r/askdocs', `r/diagnoseme', `r/medical\_advice'.

The choice for these subreddits stems from past works~\cite{mesbah-etal-2019-training,saha2019social,chancellor2019discovering}.

\section{ADR Detection Scenarios and Prompts}
\label{appendix_sec:adr_detection_prompts_criteria}

Posts on social media platforms discussing adverse drug reactions related to psychiatric medications are often written by individuals with limited or no medical knowledge. As a result, the level of certainty in expressing concerns about potential side effects can vary significantly. Some posts are more assertive, while others express uncertainty. For example, individuals may report experiencing adverse symptoms after taking psychiatric medications, be unsure if these symptoms are caused by the medication, or inquire whether their symptoms could be related to the drugs they are taking. Additionally, some posts may express concerns about possible future side effects of starting a new psychiatric medication. These scenarios were used as examples to guide both annotators and language models. At last, both LLMs and experts were asked to determine whether the concern could be related to ADR or not based on their experience. Table~\ref{tab:appendix_adr_detection_zero_shot_prompt} and Table~\ref{tab:appendix_adr_detection_few_shot_prompt} present prompts used with LLMs for detecting cases of ADR from psychiatric medication.

\begin{table*}[!htbp]
    \centering
    \small
    \begin{tabular}{p{0.10\linewidth} | p{0.85\linewidth}}
    \toprule
    \textbf{Type} & \textbf{Prompt}\\
    \toprule
         \textbf{System Prompt}&You are an expert psychiatrist and psychologist, and you help people with their mental health and adverse drug reaction (ADR) related queries. You provide advice, and guidance to people who are experiencing mental health issues. Given below are different events/ways in which concerns related to a potential case of adverse drug reaction (ADR) can be expressed (Presented as ADVERSE\_DRUG\_REACTION\_EVENTS), the Reddit post title (Presented as POST\_TITLE), and the Reddit post text (Presented as POST\_TEXT) with mentions of psychiatric medicines and symptoms.\\
         &\\
         &ADVERSE\_DRUG\_REACTION\_EVENTS: An event can potentially express concerns for an adverse drug reaction (ADR) related to psychiatric medications in one or more of the following four (4) ways:\\
         &\\
         &1) The person is experiencing some adverse symptoms after taking psychiatric medicine(s).\\
         &2) The person is unsure if the adverse symptoms are caused due to psychiatric medication(s).\\
         &3) The person is inquiring whether the adverse symptoms were potentially caused due to any psychiatric medicine taken by the person.\\
         &4) Person is not taking psychiatric medication and is concerned if taking a new psychiatric medication would have side-effects in the future.\\
         &AND\\
         &The post asks a question that is relevant to the adverse drug reaction.\\
         &\\
         &Keeping the context of the POST\_TITLE and POST\_TEXT in view and using the different ways of potential expressions provided in ADVERSE\_DRUG\_REACTION\_EVENTS, your task is to determine whether the post actually expresses about an adverse drug reaction (ADR) related to a psychiatric medication and asks a question about the ADR related to the psychiatric medication or not.\\
         &\\
         &You should think step by step and provide a rationale for your answer. You should first provide your rationale and at last you should explicitly provide a label as `ADR-Yes' or `ADR-No' determining whether the post talks about adverse drug reaction or not respectively. Exactly provide one of class label and always provide the exact class label in the format - Class Label: <ADR-Yes or ADR-No>\\
         &\\
         \midrule
         \textbf{User Prompt}&POST\_TITLE: <post\_title>\\
         &POST\_TEXT: <post\_text>\\
         \bottomrule
    \end{tabular}
    \caption{Prompt used for the ADR detection task in zero-shot setting.}
    \label{tab:appendix_adr_detection_zero_shot_prompt}
\end{table*}

\begin{table*}[!htbp]
    \centering
    \small
    \begin{tabular}{p{0.10\linewidth} | p{0.85\linewidth}}
    \toprule
    \textbf{Type} & \textbf{Prompt}\\
    \toprule
         \textbf{System Prompt}&You are an expert psychiatrist and psychologist, and you help people with their mental health and adverse drug reaction (ADR) related queries. You provide advice, and guidance to people who are experiencing mental health issues. Given below are different events/ways in which concerns related to a potential case of adverse drug reaction (ADR) can be expressed (Presented as ADVERSE\_DRUG\_REACTION\_EVENTS), 5 examples of Reddit post title, post text and class label (Presented as EXAMPLE\_POST\_TITLE, EXAMPLE\_POST\_TEXT, EXAMPLE\_CLASS\_LABEL), the Reddit post title (Presented as POST\_TITLE), and the Reddit post text (Presented as POST\_TEXT) with mentions of psychiatric medicines and symptoms.\\
         &\\
         &ADVERSE\_DRUG\_REACTION\_EVENTS: An event can potentially express concerns for an adverse drug reaction (ADR) related to psychiatric medications in one or more of the following four (4) ways:\\
         &\\
         &1) The person is experiencing some adverse symptoms after taking psychiatric medicine(s).\\
         &2) The person is unsure if the adverse symptoms are caused due to psychiatric medication(s).\\
         &3) The person is inquiring whether the adverse symptoms were potentially caused due to any psychiatric medicine taken by the person.\\
         &4) Person is not taking psychiatric medication and is concerned if taking a new psychiatric medication would have side-effects in the future.\\
         &AND\\
         &The post asks a question that is relevant to the adverse drug reaction.\\
         &\\
         &Set of Examples:\\
         &\\
         &1. EXAMPLE\_POST\_TITLE: <example\_post\_title\_1>\\
         & EXAMPLE\_POST\_TEXT: <example\_post\_text\_1>\\
         & EXAMPLE\_CLASS\_LABEL: <example\_class\_label\_1>\\
         &...\\
         &...\\
         & EXAMPLE\_CLASS\_LABEL: <example\_class\_label\_5>\\
         &\\
         &Keeping the context of the POST\_TITLE and POST\_TEXT in view and and using the 5 examples (EXAMPLE\_POST\_TITLE, EXAMPLE\_POST\_TEXT, EXAMPLE\_CLASS\_LABEL) and using the different ways of potential expressions provided in ADVERSE\_DRUG\_REACTION\_EVENTS, your task is to determine whether the post actually expresses about an adverse drug reaction (ADR) related to a psychiatric medication and asks a question about the ADR related to the psychiatric medication or not.\\
         &\\
         &You should think step by step and provide a rationale for your answer. You should first provide your rationale and at last you should explicitly provide a label as `ADR-Yes' or `ADR-No' determining whether the post talks about adverse drug reaction or not respectively. Exactly provide one of class label and always provide the exact class label in the format - Class Label: <ADR-Yes or ADR-No>\\
         &\\
         \midrule
         \textbf{User Prompt}&POST\_TITLE: <post\_title>\\
         &POST\_TEXT: <post\_text>\\
         \bottomrule
    \end{tabular}
    \caption{Prompt used for the ADR detection task in 5-shot setting.}
    \label{tab:appendix_adr_detection_few_shot_prompt}
\end{table*}

\begin{table*}[!htbp]
    \centering
    \small
    \begin{tabular}{p{0.10\linewidth} | p{0.85\linewidth}}
    \toprule
    \textbf{Type} & \textbf{Prompt}\\
    \toprule
         \textbf{System Prompt}&You are an expert psychiatrist and psychologist, and you help people with their mental health and adverse drug reaction (ADR) related queries. You provide advice, and guidance to people who are experiencing mental health issues. Given below is the list of class names and definitions for each class of adverse drug reaction (ADR) (Presented as ADR\_CLASS\_NAMES\_DEFINITION), the Reddit post title (Presented as POST\_TITLE), and the Reddit post text (Presented as POST\_TEXT) expressing adverse drug reaction (ADR) related to a psychiatric medication/s.\\
&\\
&ADR\_CLASS\_NAMES\_DEFINITION: Adverse drug reactions (ADRs) related to a psychiatric medications can be classified in one of the following classes:\\
&1) Dose-related-adr-reactions: These are the reactions that are directly related to the dosage of the psychiatric medication.\\
&2) Non-dose-adr-reactions: These are the reactions where any exposure of psychiatric medication is enough to trigger an adverse reaction.\\
&3) Dose-and-time-adr-reactions: These are the reactions that are related due to dose accumulation, or with prolonged use of the psychiatric medication.\\
&4) Time-related-adr-reactions: These are the reactions that are related due to prolonged use in a psychiatric medication which doesn't tend to accumulate.\\
&5) Withdrawal-adr-reactions: These are the reactions that are related to the undesired effects of ceasing or stopping the intake of the psychiatric medication.\\
&\\
&Keeping the context of the POST\_TITLE and POST\_TEXT in view and using the definitions provided in ADR\_CLASS\_NAMES\_DEFINITION, your task is to determine the class of adverse drug reaction (ADR) related to a psychiatric medication/s expressed in the post.\\
&\\
&You should think step by step and provide a rationale for your answer. You should first provide your rationale and at last you should explicitly provide the class label from ADR\_CLASS\_NAMES\_DEFINITION which is most appropriate and applicable for the post. Only provide one of class label and always provide the exact class label in the format - Class Label: <label name>\\
         &\\
         \midrule
         \textbf{User Prompt}&POST\_TITLE: <post\_title>\\

         &POST\_TEXT: <post\_text>\\
         \bottomrule
    \end{tabular}
    \caption{Prompt used for the ADR multiclass classification task in zero-shot setting.}
    \label{tab:appendix_adr_subclassification_zero_shot_prompt}
\end{table*}

\begin{table*}[!htbp]
    \centering
    \small
    \begin{tabular}{p{0.10\linewidth} | p{0.85\linewidth}}
    \toprule
    \textbf{Type} & \textbf{Prompt}\\
    \toprule
         \textbf{System Prompt}&You are an expert psychiatrist and psychologist, and you help people with their mental health and adverse drug reaction (ADR) related queries. You provide advice, and guidance to people who are experiencing mental health issues. Given below is the list of class names and definitions for each class of adverse drug reaction (ADR) (Presented as ADR\_CLASS\_NAMES\_DEFINITION), 2 examples of reddit post title, post text and class label (Presented as EXAMPLE\_POST\_TITLE, EXAMPLE\_POST\_TEXT, EXAMPLE\_CLASS\_LABEL), the Reddit post title (Presented as POST\_TITLE), and the Reddit post text (Presented as POST\_TEXT) expressing adverse drug reaction (ADR) related to a psychiatric medication/s.\\
&\\
&ADR\_CLASS\_NAMES\_DEFINITION: Adverse drug reactions (ADRs) related to a psychiatric medications can be classified in one of the following classes:\\
&1) Dose-related-adr-reactions: These are the reactions that are directly related to the dosage of the psychiatric medication.\\
&2) Non-dose-adr-reactions: These are the reactions where any exposure of psychiatric medication is enough to trigger an adverse reaction.\\
&3) Dose-and-time-adr-reactions: These are the reactions that are related due to dose accumulation, or with prolonged use of the psychiatric medication.\\
&4) Time-related-adr-reactions: These are the reactions that are related due to prolonged use in a psychiatric medication which doesn't tend to accumulate.\\
&5) Withdrawal-adr-reactions: These are the reactions that are related to the undesired effects of ceasing or stopping the intake of the psychiatric medication.\\
&\\
&Set of Examples:\\
&\\
&1) EXAMPLE\_POST\_TITLE: <example\_title\_1>\\
&EXAMPLE\_POST\_TEXT: <example\_text\_1>\\
&EXAMPLE\_CLASS\_LABEL: <example\_class\_label\_1>\\
&...\\
&EXAMPLE\_CLASS\_LABEL: <example\_class\_label\_5>\\
&\\
&Keeping the context of the POST\_TITLE and POST\_TEXT in view and using the 2 examples (EXAMPLE\_POST\_TITLE, EXAMPLE\_POST\_TEXT, EXAMPLE\_CLASS\_LABEL) and definitions provided in ADR\_CLASS\_NAMES\_DEFINITION, your task is to determine the class of adverse drug reaction (ADR) related to a psychiatric medication/s expressed in the post. You should think step by step and provide a rationale for your answer. You should first provide your rationale and at last you should explicitly provide the class label from ADR\_CLASS\_NAMES\_DEFINITION which is most appropriate and applicable for the post. Only provide one of class label and always provide the exact class label in the format - Class Label: <label name>\\

         &\\
         \midrule
         \textbf{User Prompt}&POST\_TITLE: <post\_title>\\
         &POST\_TEXT: <post\_text>\\
         \bottomrule
    \end{tabular}
    \caption{Prompt used for the ADR multiclass classification task in 5-shot setting.}
    \label{tab:appendix_adr_subclassification_few_shot_prompt}
\end{table*}

\section{ADR Multiclass Classification Definitions and Prompts}
\label{appendix_sec:adr_multiclass_definitions}

We provided the same definitions to both the LLMs and expert annotators for the annotation task. To independently evaluate the LLMs, we focused only on posts annotated as expressing ADR-related concerns ($N=$133) in the \dataset~benchmark. Adverse drug reactions (ADRs) related to a psychiatric medications can be classified in one of the following classes:

\begin{itemize}
    \item \textbf{Dose-related}: These are the reactions that are directly related to the dosage of the psychiatric medication.\\
    \vspace{-2em}
    \item \textbf{Non-dose-related}: These are the reactions where any exposure of psychiatric medication is enough to trigger an adverse reaction.\\
    \vspace{-2em}
    \item \textbf{Time-related}: These are the reactions that are related due to prolonged use in a psychiatric medication which doesn't tend to accumulate.\\
    \vspace{-2em}
    \item \textbf{Dose-and-time-related}: These are the reactions that are related due to dose accumulation, or with prolonged use of the psychiatric medication.\\
    \vspace{-2em}
    \item \textbf{Withdrawal}: These are the reactions that are related to the undesired effects of ceasing or stopping the intake of the psychiatric medication.
\end{itemize}

Table~\ref{tab:appendix_adr_subclassification_zero_shot_prompt} and~\ref{tab:appendix_adr_subclassification_few_shot_prompt} present the LLM prompts used for zero- and few-shot ADR multiclass classification. 

\section{Annotation Task Details}
\label{appendix_sec:expert_annotation_task_details}

\begin{figure}[!h]
    \centering
    \small
    \includegraphics[width=\columnwidth]{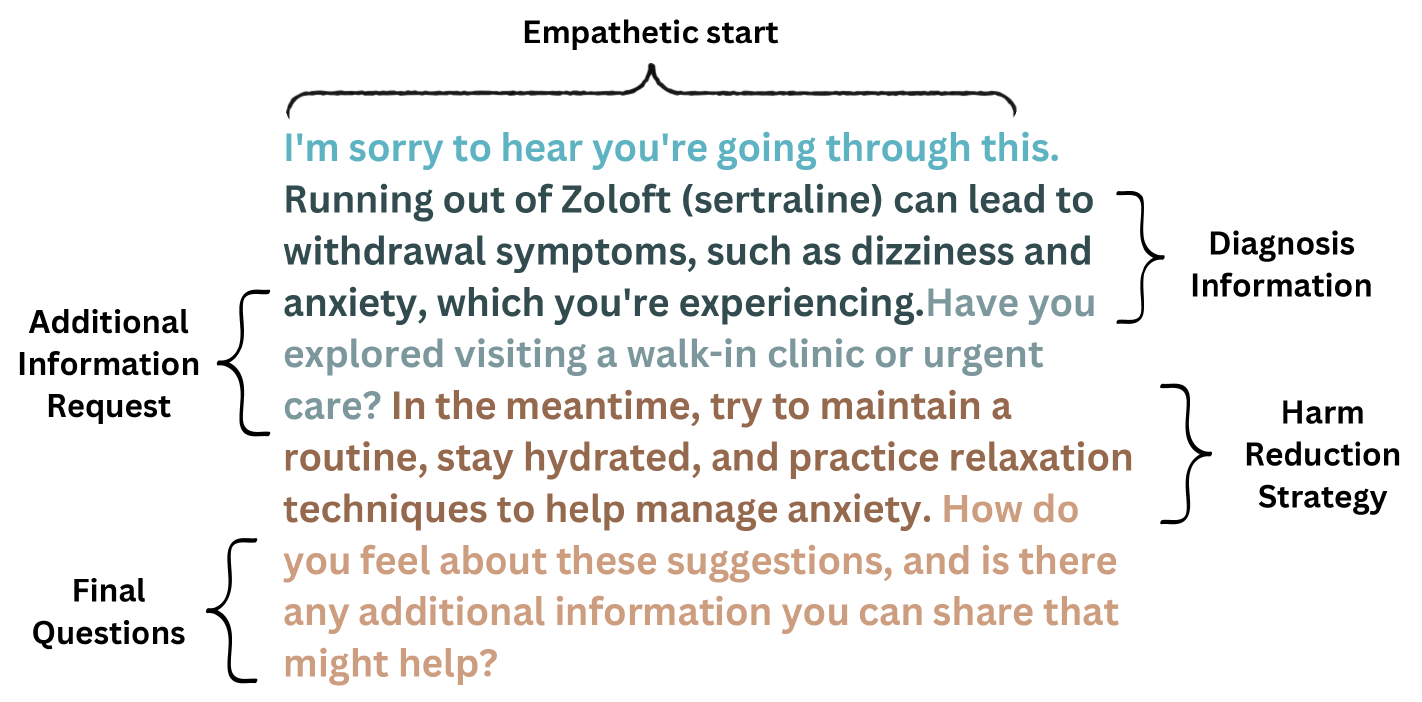}
    \caption{Sample answer representing the structure of answers provided in the~\dataset~benchmark dataset.}
    \label{fig:appendix_dataset_answer_writing_sample}
\end{figure}

We collaborated with a team of four medical experts (three doctors and one medical student), all of whom are co-authors of this work. Hence, we did not provide any additional compensation for the annotation task. Furthermore, Institutional Review Board (IRB) approval was obtained before the annotation task. To facilitate the annotation process, we developed a custom web-based tool specifically for annotating the~\dataset~benchmark. Figure~\ref{fig:appendix_annotation_platform_screenshots} presents the interface of the annotation tool used for the data annotation purpose. We conducted a preliminary round of test annotations to familiarize the annotators with both the criteria and the annotation tool. For the further rounds, the average Fleiss' kappa inter-annotator agreement was $\sim \kappa=$0.33, with all three annotators agreeing on the labels for 48\% of the posts, indicating a fair level of agreement~\cite{landis1977measurement}. These results are consistent with previous research, which has reported similar inter-annotator agreement scores for tasks of comparable difficulty~\cite{karpinska2021perils, saha2021short}.

\subsection{ADR Post Reply Annotation and Generation}

Figure~\ref{fig:appendix_dataset_answer_writing_sample} presents the sample of structure of the expert responses provided in the~\dataset~benchmark. The most experienced doctor on the collaborating team provided the responses, with each taking $\sim$8 minutes to answer on average. We provided the same set of instructions to the domain expert and the LLMs for writing the response to an ADR related query. Both the experts and the LLMs were prompted to write responses that follow a strict logical structure typically seen in clinical settings:

\begin{itemize}
    \item Start with empathizing with the post writer and acknowledging their concern.\\
    \item As step 2, provide a brief explanation for the potential diagnosis as an adverse drug reaction.\\
    \item As step 3, talk briefly about the evidence of the adverse drug reaction which is available in the post and provide a brief reasoning for proposing a harm reduction strategy.\\
    \item As step 4, ask potential questions to get additional information that might help in providing a better response.\\
    \item As step 5, provide a brief explanation of the harm reduction strategy that the poster should follow to get relief.\\
    \item Finally, ask the post writer about how they feel about the proposed harm reduction strategy.  
\end{itemize}

Table~\ref{tab:appendix_adr_reply_generation_prompt} presents the prompt for ADR reply prompt generation.

\begin{table*}[!htbp]
    \centering
    \small
    \begin{tabular}{p{0.10\linewidth} | p{0.85\linewidth}}
    \toprule
    \textbf{Type} & \textbf{Prompt}\\
    \toprule
         \textbf{System Prompt}&You are an expert psychiatrist and psychologist, and you help people with their mental health and adverse drug reaction (ADR) related queries. You provide advice, and guidance to people who are experiencing mental health issues. Given below is the guideline for generating an ideal reply (Presented as IDEAL\_REPLY\_TEMPLATE), the reddit post title (Presented as POST\_TITLE), and the reddit post text (Presented as POST\_TEXT) expressing adverse drug reaction (ADR) related to a psychiatric medication/s.\\
&\\
&IDEAL\_REPLY\_TEMPLATE: An "ideal reply" follows the below steps:\\
&1. Start with empathizing with the post writer and acknowledging their concern.\\
&2. As step 2, provide a brief explanation for the potential diagnosis as an adverse drug reaction.\\
&3. As step 3, talk briefly about the evidence of the adverse drug reaction which is available in the post and provide a brief reasoning for proposing a harm reduction strategy.\\
&4. As step 4, ask potential questions to get additional information that might help in providing a better response.\\
&5. As step 5, provide a brief explanation of the harm reduction strategy that the poster should follow to get relief.\\
&6. Finally, ask the post writer about how they feel about the proposed harm reduction strategy.\\
&\\    
&Below are some additional guidelines:\\
&1. Do not words like dear poster, dear user, best regards, etc. in the response.\\
&2. Some of the steps of an "ideal answer" are optional and you can skip those if that makes more sense, but it is recommended to include them.\\
&3. The response should be concise and to the point and with the word limit of 225 words or 300 token. Never exceed this word/token limit.\\
&\\
&Keeping the context of the POST\_TITLE and POST\_TEXT in view and using the guideline provided in IDEAL\_REPLY\_TEMPLATE, your task is to generate a reply to the post. Your response should be helpful and aim to provide a solution to the issues/problems mentioned in the post.\\

         &\\
         \midrule
         \textbf{User Prompt}&POST\_TITLE: <post\_title>\\
         &POST\_TEXT: <post\_text>\\
         \bottomrule
    \end{tabular}
    \caption{Prompt used for generating the reply for post expressing ADR related concerns.}
    \label{tab:appendix_adr_reply_generation_prompt}
\end{table*}

\section{Model Details, Hyperparameters, and Compute}
\label{appendix_sec:compute_details}

We use API-based model inference for GPT4-Turbo, GPT-4o, Llama 3.1-70B Instruct, Llama 3.1-405B Instruct, Claude 3 Haiku, Claude 3 Opus, Claude 3.5 Sonnet. We used Azure OpenAI service for accessing GPT4-Turbo \& GPT-4o models, Together.ai for accessing Llama 3.1-70B Instruct \& Llama 3.1-405B Instruct, and the Anthropic platform for accessing the Claude series models. For OpenBioLLM-Llama3-70B and Llama3-Med42v2-70B, we did GPU-based inference using an 8x NVIDIA L40S GPU cluster. The hyperparameters for API-based inference models and GPU-based inference models are presented below. Table~\ref{tab:appendx_model_details} presents the details regarding the model sizes and versions.
\\

\noindent\textbf{Hyperparameters for GPT, Llama 3.1 and Claude series models}: temperature $t=0$ (for ADR detection and multiclass classification) \& $t=0.6$ (for response generation), top\_p$=1$, frequency\_penalty$=0$, presence\_penalty$=0$, max\_tokens$=600$ (for ADR detection and multiclass classification) \& max\_tokens$=340$ (for response generation), number of completions$=1$, top\_k$=50$ (for Claude series models)\\

\noindent\textbf{Hyperparameters for OpenBioLLM-Llama3-70B and Llama3-Med42v2-70B}: temperature $t=0$ (for ADR detection and multiclass classification) \& $t=0.6$ (for response generation), top\_p$=1$, frequency\_penalty$=0$, presence\_penalty$=0$, max\_tokens$=600$ (for ADR detection and multiclass classification) \& max\_tokens$=340$ (for response generation), number of completions$=1$, top\_k$=50$. 

\section{ADR detection and multiclass classification results}
\label{appendix_sec:rq_1_additional_results}

\begin{figure*}[!h]
    \centering
    \includegraphics[width=\linewidth]{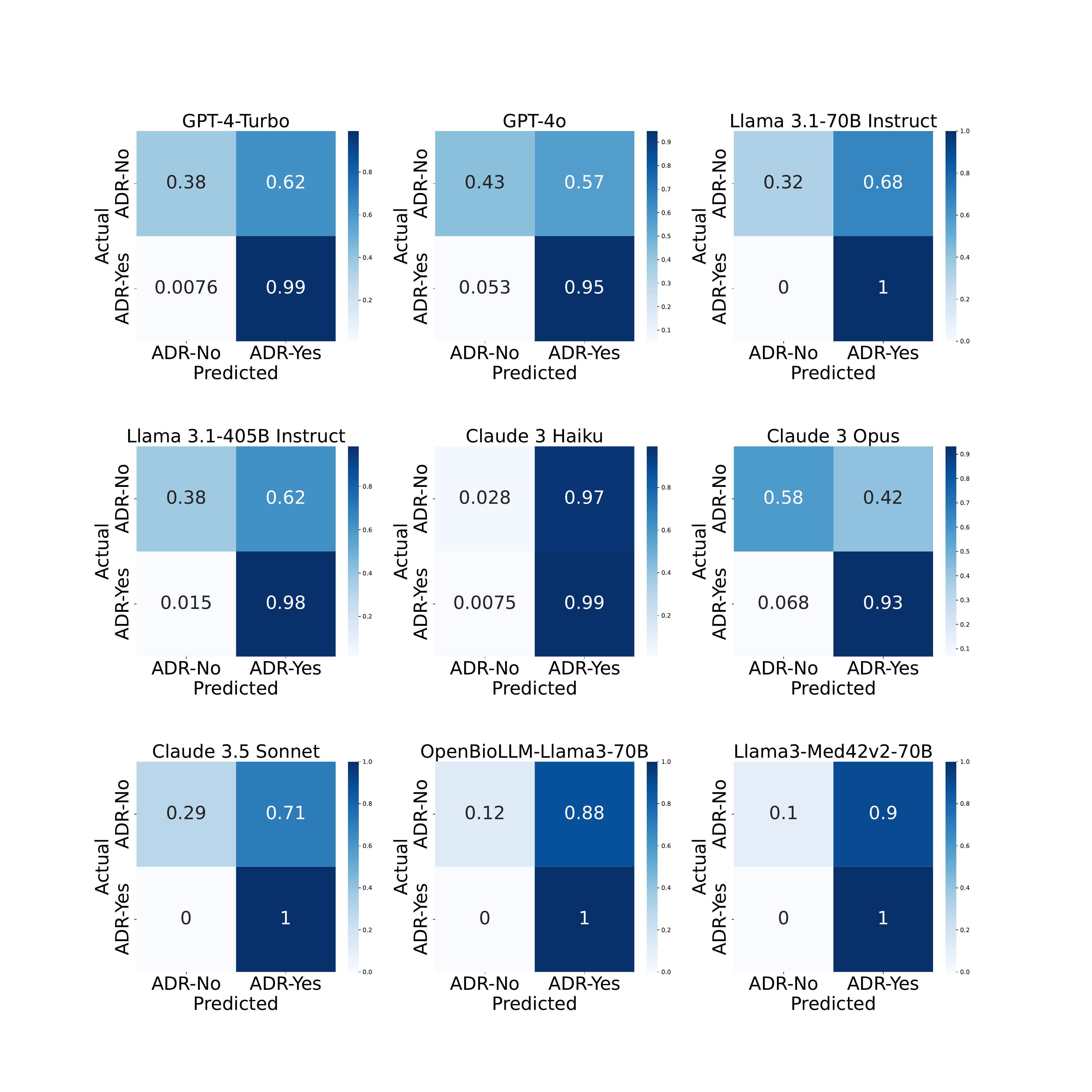}
    \caption{Confusion Matrix for ADR Detection task in zero-shot setting. The values represents the ratio of examples in the predicted class over total number of examples in the actual class.}
    \label{fig:appendix_adr_binary_detection_cm}
\end{figure*}

\begin{figure*}[!h]
    \centering
    \includegraphics[width=\linewidth]{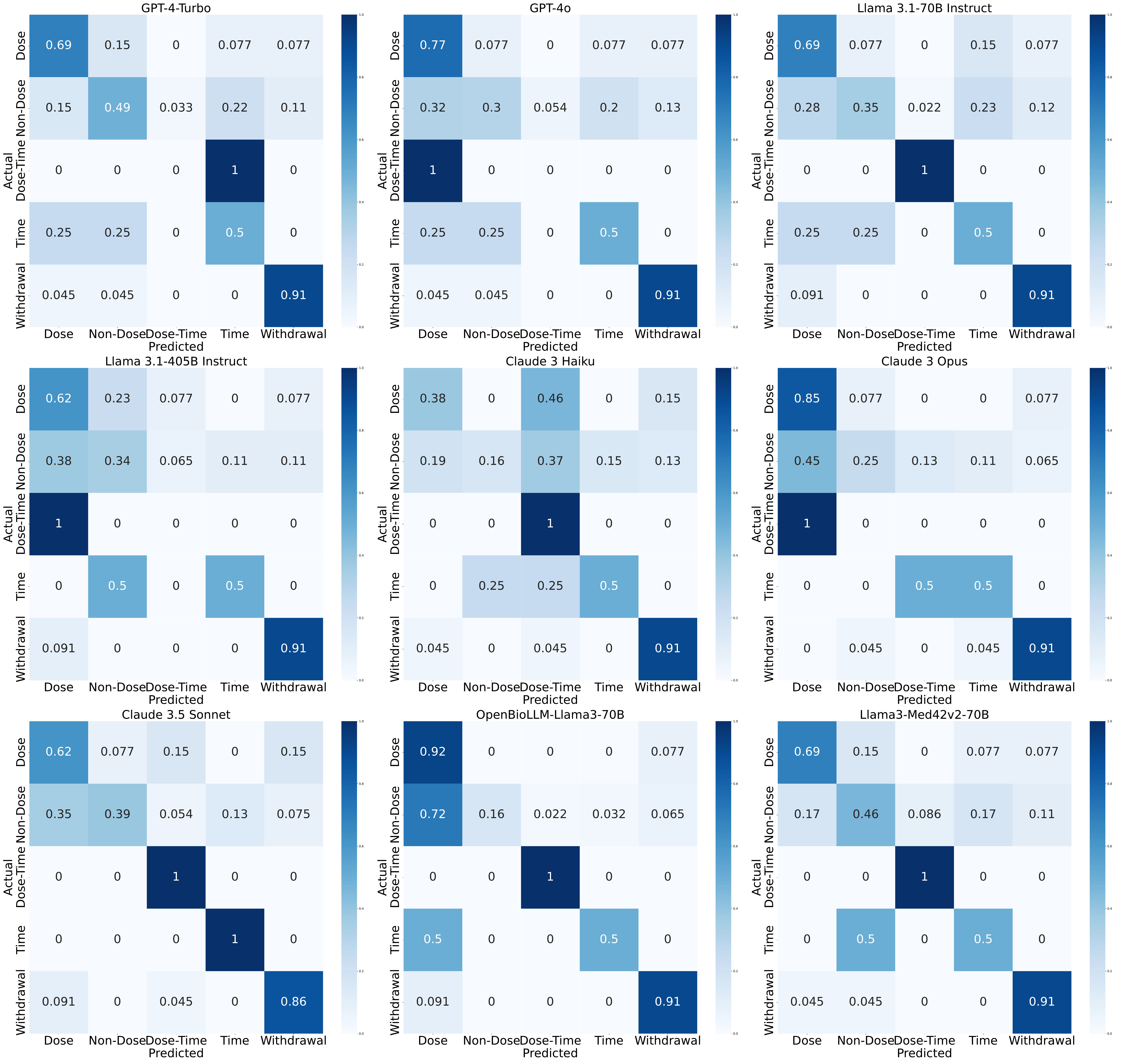}
    \caption{Confusion Matrix for ADR Multiclass classification task in zero-shot setting. The values represents the ratio of number of examples in the predicted class over the total number of examples in the actual (ground truth) class.}
    \label{fig:appendix_adr_multiclass_zero_shot_cm}
\end{figure*}

We analyzed the class-wise distribution of predicted labels for the ADR detection and ADR multiclass classification task. Figure~\ref{fig:appendix_adr_binary_detection_cm} and Figure~\ref{fig:appendix_adr_multiclass_zero_shot_cm} present the confusion matrices for the ADR detection and ADR multiclass classification task in zero-shot setting. Analyzing Figure~\ref{fig:appendix_adr_binary_detection_cm} we observed that all models performed exceptionally well in cases of ADRs with 4 out of 9 models correctly detecting all examples in the `ADR-Yes' class. However, all models struggled in correctly classifying cases of `ADR-No' class with the best model (Claude 3 Opus) misclassifying 42\% examples. This qualitatively implied that models showed lack of lived experience and a ``risk-averse'' behavior. Analyzing the ADR multiclass classification results in zero-shot setting (Figure~\ref{fig:appendix_adr_multiclass_zero_shot_cm}), we observed that \textit{Withdrawal} ADRs were correctly classified more than 90\% times by all models. However, all LLMs struggled between the \textit{Dose} and \textit{Non-Dose} related ADRs and failed to understand the nuances between the two types of ADRs.

\section{ADR Detection and Multiclass Classification Error Analysis}
\label{appendix_sec:rq_1_adr_detection_error_analysis}

We conducted a qualitative error analysis for misclassified examples in the ADR detection and multiclass classification tasks, focusing on Claude 3 Opus and Llama 3.1 405B in few-shot settings. Upon the analysis, two major themes emerged: (a) lack of lived experience and (b) incorrect assumptions about potential ADR queries. In the first set of errors, models adhered too rigidly to prompt rules, missing other possible symptom explanations. In the second set of errors, models often confused posts seeking emotional support with ADR-related queries. The model misjudged a person's use of social media to share their feelings as a potential ADR-related query.

For cases where the model demonstrates a lack of lived experience, we observe expert quotes such as \textit{``Patient is having swallowing difficulties which seems to be due to GI issues rather than medication''} and \textit{``We can not say she has an ADR since she is actually sleep deprived, plus slightly (minimally) overweight, so we should need to assess if she actually has sleep apnea.''}. These quotes indicate that the model is quick to label a post as ADR and can overlook some other contributing factors for the symptoms, while the experts are cautious while labeling a post as ADR. Getting a diagnosis of ADR by psychiatric medication can be overwhelming for patients already suffering from anxiety, depression, and other ailments, and eliminating other potential causes first is a smarter approach.

Reddit is a social media space where people not only ask queries but often share their feelings and thoughts. The model confuses posts of people sharing what they are going through and their experiences as people seeking ADR-related help even if no explicit question has been asked. There are also cases where the question being asked in the post is about the workings of a particular drug, lifestyle, or something else unrelated to ADR but the model gets confused. Experts are able to identify these correctly and give valid reasoning such as \textit{``Although the post has a detailed description of an ADR (Serotonin syndrome), the patient doesn't have any explicit questions and instead just seems to be sharing her situation.''} and \textit{``No ADR, just questions on how the drug works.''} for these cases.

\section{Methodological Details for Emotional and Tone Alignment}
\label{sec:appendix_rq2_emotional_tone_alignment}

To compute the emotional and tonal alignment, we lemmatized the Empath lexicon, expert responses and LLM responses using `en\_core\_web\_sm' model on SpaCy~\cite{ines_montani_2023_10009823}. This preprocessing step ensured consistency in comparing the linguistic features across the responses with the Empath categories.

\begin{figure*}
    \centering
    \includegraphics[width=\textwidth]{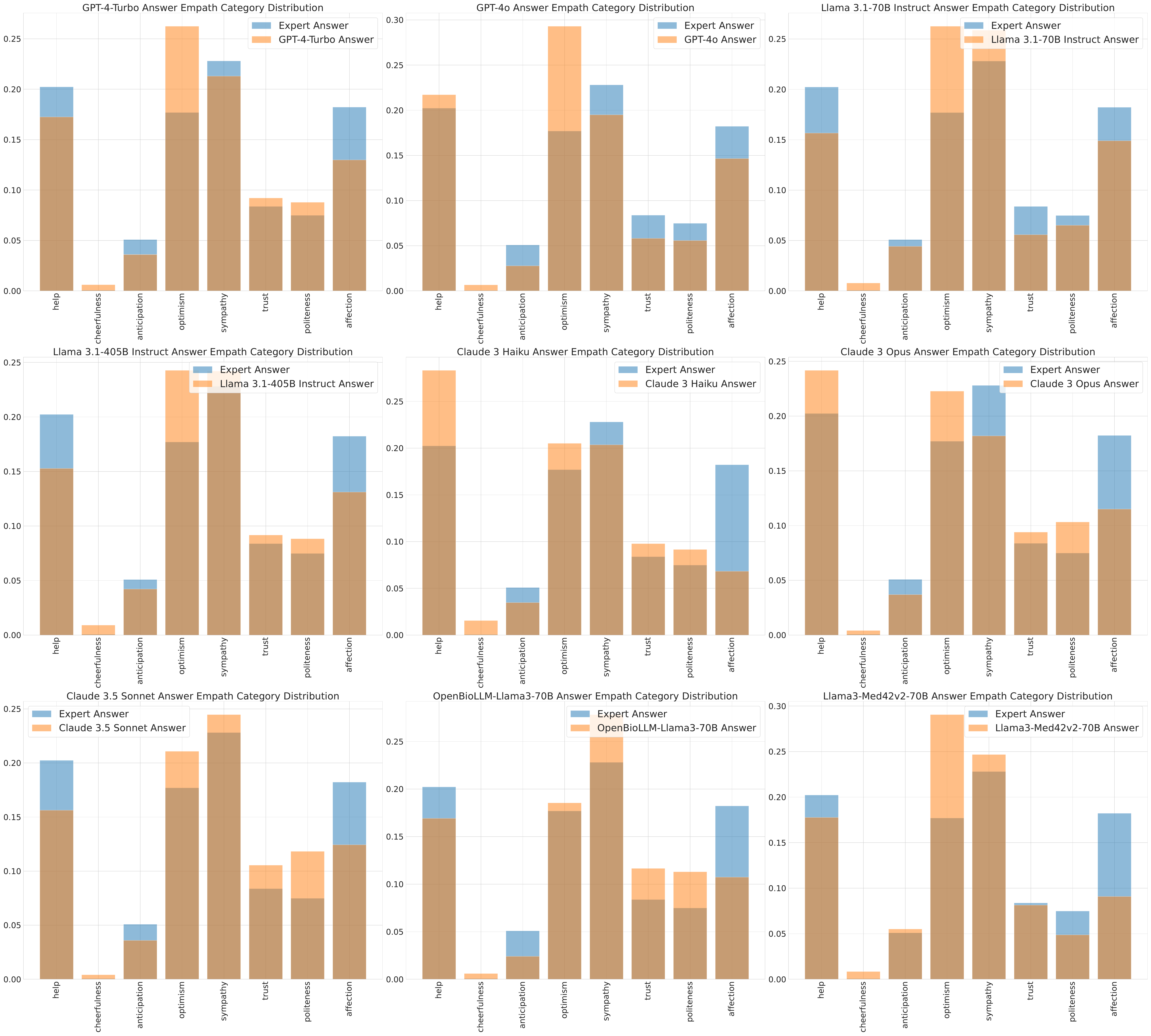}
    \caption{Average Distribution of Emotion and Tone categories from Empath across LLM and Expert responses.}
    \label{fig:appendix_empath_category_distribution}
\end{figure*}

\section{Harm Reduction Strategy Alignment Qualitative Analysis}
\label{appendix_sec:hrs_alignment_error_analysis}

We qualitatively analyzed alignment between the LLM's harm reduction strategies and those suggested by the expert. One pattern observed across all LLMs was that in addition to their main response to the issue, they suggested non-pharmacological advice on lifestyle changes involving sleep hygiene (\textit{``Prioritize adequate sleep.''}), diet (\textit{``Maintain a balanced diet.''}) and mindfulness techniques such as meditation (\textit{``Practice stress-management techniques like deep breathing.''}) and journaling (\textit{``Keep a journal of your symptoms.''}). On the other hand, expert answers tended to focus on addressing the symptoms or questions involving the medication. Harm reduction strategies suggested by LLMs related to medication were often paired with an action to discuss with a doctor about the recommendations before committing to them (\textit{``Consider adjusting the dosage with your doctor's guidance.''}, \textit{``Gradually taper off Trintellix under medical supervision.''}). Examples of alignment in these cases are presented in Table \ref{tab:hrs_alignment_examples}.

\begin{table*}[!h]
\centering
\begin{tabular}{|p{0.5\textwidth}|p{0.5\textwidth}|}
    \hline
    \vspace{1pt}\thead{\textbf{\normalsize{Atomic Strategies}}} & \vspace{1pt}\thead{\textbf{\normalsize{Combined Strategies}}} \\
    
    \hline
    \makecell[l]{\minipage{0.5\textwidth} \vspace{5pt}\small{\textcolor{teal}{1. Consult your healthcare provider before altering your medication dosage.} \\ \textcolor{teal}{2. Discuss symptoms and tapering plans with your doctor.} \\ 3. Avoid activities that could be dangerous due to vision issues, such as driving.}\endminipage}
    & 
    \makecell[l]{\minipage{0.5\textwidth}\vspace{5pt}\small{\textcolor{teal}{1. Consult your healthcare provider before altering your medication dosage. Discuss symptoms and tapering plans with your doctor.}\\2. Avoid activities that could be dangerous due to vision issues, such as driving.}\endminipage}
    \\
    \hline
    \makecell[l]{\minipage{0.5\textwidth}\vspace{5pt}\small{\textcolor{teal}{1. Taper off the medication very gradually.\\2. Extend the tapering period.\\3. Make smaller dosage reductions.}\\4. Discuss the possibility of using other supportive medications or therapies with your healthcare provider.}\endminipage}
    & 
    \makecell[l]{\minipage{0.5\textwidth}\vspace{5pt}\small{\textcolor{teal}{1. Taper off the medication very gradually. Extend the tapering period. Make smaller dosage reductions.}\\2. Discuss the possibility of using other supportive medications or therapies with your healthcare provider.}\endminipage}
    \\
    \hline
    \makecell[l]{\minipage{0.5\textwidth} \vspace{5pt}\small{1. Discuss medication concerns with your psychiatrist.\\\textcolor{teal}{2. Gradually switch to another antidepressant under supervision.\\3. Consider alternative antidepressants like bupropion.}}\endminipage}
    & 
    \makecell[l]{\minipage{0.5\textwidth}\vspace{5pt}\small{1. Discuss medication concerns with your psychiatrist.\\\textcolor{teal}{2. Gradually switch to another antidepressant under supervision. Consider alternative antidepressants like bupropion.}}\endminipage}
    \\    
    \hline
    \makecell[l]{\minipage{0.5\textwidth} \vspace{5pt}\small{\textcolor{teal}{1. Review the timing and dosage of medications under the guidance of a psychiatric practitioner.\\2. Adjust the time you take Adderall XR.}\\ \textcolor{brown}{3. Maintain a sleep routine.\\4. Use a sleep mask.}\\5. Discuss with your psychiatrist the possibility of using a different sleep aid.\\\textcolor{teal}{6. Consider adjusting the Lamotragine dosage if it's found to be the cause.}}\endminipage}
    & 
    \makecell[l]{\minipage{0.5\textwidth}\vspace{5pt}\small{\textcolor{teal}{1. Review the timing and dosage of medications under the guidance of a psychiatric practitioner. Adjust the time you take Adderall XR. Consider adjusting the Lamotragine dosage if it's found to be the cause.}\\\textcolor{brown}{2. Maintain a sleep routine. Use a sleep mask.}\\3. Discuss with your psychiatrist the possibility of using a different sleep aid.}\endminipage}
    \\      
    \hline
    \makecell[l]{\minipage{0.5\textwidth} \vspace{5pt}\small{1. Switch to another medication if needed.\\\textcolor{teal}{2. Practice mindfulness techniques.\\3. Use relaxation techniques to manage restlessness.}}\endminipage}
    & 
    \makecell[l]{\minipage{0.5\textwidth}\vspace{5pt}\small{1. Switch to another medication if needed.\textcolor{teal}{\\2. Practice mindfulness techniques. Use relaxation techniques to manage restlessness.}}\endminipage}
    \\      
    \hline
\end{tabular}
\caption{Examples of harm reduction strategies that were combined by GPT-4o (groups highlighted in different colors). The combination was performed for strategies which suggest the same overall approach with minor differences in specific details.}
\label{tab:hrs_combination_examples}
\end{table*}

\begin{table*}[h]
\centering
\begin{tabular}{|p{0.7\textwidth}|p{0.3\textwidth}|}
    \hline
    \thead{\textbf{\normalsize{Expert's Response}}} & \thead{\textbf{\normalsize{LLM's Harm}}\\\textbf{\normalsize{Reduction Strategies}}} \\
    
    \hline
    \makecell[l]{\minipage{0.7\textwidth} \vspace{5pt}\small{\textit{"First of all, I understand how difficult living with this experience ... It is always important to make all medication adjustment with close monitorization from your main provider. Therefore, \textcolor{teal}{the main recommendation would be to seek for professional help, either by making an appointment with your main provider (if possible) or going to a psychiatry emergency department (if symptoms become severe)}. There are many strategies for tapering antidepressant medication ...}''}\endminipage}
    & 
    \makecell[l]{\minipage{0.3\textwidth}\vspace{5pt}\small{\textcolor{teal}{1. Consult a healthcare professional familiar with psychiatric medications and withdrawal symptoms.}\\ \textcolor{amaranth}{2. Prioritize adequate sleep. Focus on nutrition.\\3. Implement stress management techniques.}}\endminipage}
    \\ 
    \hline
    \makecell[l]{\minipage{0.7\textwidth} \vspace{5pt}\small{\textit{"I understand how frustrating living with this feeling might be ... It always depends on each patient's perception and feelings about it. How is your own experience about it? Do you think that it is being helpful or is it worsening your healing process? \textcolor{teal}{If you are having difficulties to handle with it we could switch to a different antidepressant to try to avoid this side effect.} How do you feel with this? Do you feel more confident with any ... }''}\endminipage}
    &
    \makecell[l]{\minipage{0.3\textwidth}\vspace{5pt}\small{\textcolor{amaranth}{1. Discuss with your doctor about adjusting the dosage.}\\ \textcolor{teal}{2. Consider alternative medications.}\\\textcolor{amaranth}{3. Combine medication with therapy, such as cognitive-behavioral therapy.}}\endminipage}
    \\
    \hline
\end{tabular}
\caption{Examples of harm reduction strategies from LLMs and their alignment with the expert's response (aligned strategies are shown in \textcolor{teal}{teal}, non-aligned strategies in \textcolor{amaranth}{red}).}
\label{tab:hrs_alignment_examples}
\end{table*}

\subsection{Human Evaluation on LLM-based tasks for HRS}
\label{appendix_sec:hrs_annotation_correlation}

For correlation on HRS extraction and alignment between LLMs and humans, we reported the percentage of HRS where the annotator agreed with the LLM's extraction and alignment classification. For combination, since there could be multiple groups formed from different HRS, we reported the percentage of answers where the annotator agreed with the combined HRS that were generated.  

\begin{table*}[!htbp]
    \centering
    \small
    \begin{tabular}{p{0.10\linewidth} | p{0.85\linewidth}}
    \toprule
    \textbf{Type} & \textbf{Prompt}\\
    \toprule
         \textbf{System Prompt}&A harm reduction strategy is defined as a measure to be taken by an individual to reduce the negative effects of consuming a psychiatric medication. This could include changing the dosage (frequency or time of taking it) of a medication, doing exercises, avoiding certain food items, taking alternative medication or treatment, consulting a healthcare provider etc.\\
&\\
&Instructions:\\
&1. You are given a RESPONSE from a health expert. Your task is to extract as a list of atomic harm reduction strategies from the RESPONSE.\\
&2. An atomic harm reduction strategy should contain an action verb and contain a single piece of advice.\\
&3. An atomic harm reduction strategy should be extracted from a statement in the RESPONSE and not from a question.\\
&4. Each atomic harm reduction strategy should carry an entirely different piece of advice, and should be independent of other atomic harm reduction strategies in the list.\\
&5. You should only output the atomic harm reduction strategies as a list, with each item starting with "- ". Do not include other formatting.\\
         &\\
         \midrule
         \textbf{User Prompt}&RESPONSE: <response>\\
         \bottomrule
    \end{tabular}
    \caption{Prompt used for the harm reduction strategy extraction task.}
    \label{tab:appendix_hrs_extraction_prompt}
\end{table*}

\begin{table*}[!htbp]
    \centering
    \small
    \begin{tabular}{p{0.10\linewidth} | p{0.85\linewidth}}
    \toprule
    \textbf{Type} & \textbf{Prompt}\\
    \toprule
         \textbf{System Prompt}&A harm reduction strategy is defined as a measure to be taken by an individual to reduce the negative effects of consuming a psychiatric medication. This could include changing the dosage (frequency or time of taking it) of a medication, doing exercises, avoiding certain food items, taking alternative medication or treatment, consulting a healthcare provider etc.\\
&\\
&Instructions:\\
&1. You are given a list of harm reduction strategies from a health expert.\\
&2. Your task is to combine similar harm reduction strategies by logically grouping them based on the similarity of the advice.\\
&3. Two harm reduction strategies are similar if they suggest the same overall approach with differences only in the specific details.\\
&4. Do not alter the wording of any of the harm reduction strategies, only group them as multiple sentences in a single combined harm reduction strategy.\\
&5. You should only output the combined harm reduction strategies as a list, with each item starting with "- ". Do not include other formatting.\\
&\\
&You should combine strategies based on groups such as:\\
&1. Lifestyle changes such as sleeping patterns, diet, physical exercise.\\
&2. Mindfulness-based exercises.\\
&3. Adjusting dosage of existing medication.\\
&4. Trying out new medications.\\
&5. Consulting people for different opinions.\\
&\\
&You should NOT:\\
&1. Combine strategies purely based on the person involved in the suggestion (e.g: doctor).\\
&2. Combine strategies that suggest full-fledged therapy approaches with those that suggest simple self-imposed mindfulness exercises.\\
&\\
&Do this for the harm reduction strategies under "Your Task:".\\
&\\
&Consider the following examples:\\
&\\
&Harm Reduction Strategies:\\
&<example\_1\_harm\_reduction\_strategies\_list>\\
&Reasoning:\\
&<example\_1\_reasoning>\\
&Combined Harm Reduction Strategies:\\
&<example\_1\_combined\_harm\_reduction\_strategies\_list>\\
&...\\
&...\\
&<example\_5\_combined\_harm\_reduction\_strategies\_list>\\
         &\\
         \midrule
         \textbf{User Prompt}&Your Task:\\
         &Harm Reduction Strategies:\\
         &<harm\_reduction\_strategies\_list>\\
         \bottomrule
    \end{tabular}
    \caption{Prompt used for the harm reduction strategy combination task in 5-shot setting.}
    \label{tab:appendix_hrs_combination_prompt}
\end{table*}

\begin{table*}[!htbp]
    \centering
    \small
    \begin{tabular}{p{0.10\linewidth} | p{0.85\linewidth}}
    \toprule
    \textbf{Type} & \textbf{Prompt}\\
    \toprule
         \textbf{System Prompt}&You are an intelligent agent who is given a RESPONSE from a psychiatrist to a patient, and a LIST OF STATEMENTS that are harm reduction strategies. A STATEMENT is considered 'Suggestion-Present' if it can be broadly inferred implicitly OR explicitly as a harm reduction strategy suggested in the RESPONSE, or else it is 'Suggestion-NotPresent'. A 'Suggestion-Present' STATEMENT can be a specific instantiation of a broad harm reduction strategy mentioned in the RESPONSE or vice-versa.\\
&\\
&The RESPONSE may contain generic names of medication, while a STATEMENT may use a brand name for the same medication, note that these are considered the SAME.\\
&\\
&Instructions:\\
&1. The following LIST OF STATEMENTS is related to the context of the given RESPONSE.\\
&2. Your task is to analyze if EACH STATEMENT is considered 'Suggestion-Present' or 'Suggestion-NotPresent', based on the given definition and the RESPONSE.\\
&3. One by one, for each STATEMENT, mention step-by-step reasoning behind the classification, along with the label. The reasoning and classification for each STATEMENT should be independent of other STATEMENTS.\\
&4. After doing this for each STATEMENT, state the total NUMBER of 'Suggestion-Present' STATEMENTS, in a new line starting with "Number of 'Suggestion-Present' statements in total:".\\
&5. Answer ONLY in plain text (without Markdown formatting) for the RESPONSE and LIST OF STATEMENTS under "Your Task".\\
&\\
&Consider the following examples:\\
&\\
&RESPONSE: <example\_1\_response>\\
&LIST OF STATEMENTS:\\
&<example\_1\_harm\_reduction\_strategies\_list>\\
&SOLUTION:\\
&<example\_1\_solution>\\
&...\\
&...\\
&<example\_5\_solution>\\
         &\\
         \midrule
         \textbf{User Prompt}&Your Task:\\
         &RESPONSE: <response>\\
         &\\
         &LIST OF STATEMENTS:\\
         &<harm\_reduction\_strategies\_list>\\
         \bottomrule
    \end{tabular}
    \caption{Prompt used for the harm reduction strategy alignment task in 5-shot setting.}
    \label{tab:appendix_hrs_alignment_prompt}
\end{table*}

\section{Actionability Criteria}
\label{sec:appendix_actionability_criteria}

We present the concrete definitions for each of the sub-dimensions of actionability. \begin{itemize}
    \item \textbf{Practicality:} The proposed strategy should clearly identify at least one action the user can take. Further, it should be contextually feasible/practical, considering their personal circumstances, such as physical ability, financial resources, and time constraints.
    \item \textbf{Contextual relevance:} The provided strategy should be relevant and should contribute to addressing the concern of the patient.
    \item \textbf{Specificity:} The details and instructions provided in the harm reduction strategy should not be vague and should leave no to little room for risky interpretation.
    \item \textbf{Clarity:} The strategy provided can be broken up into clear, manageable and explicit steps.
\end{itemize}
Table \ref{tab:appendix_hrs_actionability_prompt} presents the LLM prompts used to get the actionability score for each harm reduction strategy.
\begin{table*}[!htbp]
    \centering
    \small
    \begin{tabular}{p{0.10\linewidth} | p{0.85\linewidth}}
    \toprule
    \textbf{Type} & \textbf{Prompt}\\
    \toprule
         \textbf{System Prompt}&You are an expert psychiatrist and psychologist. You are also an expert in identifying the practicality, contextual relevance, specificity, and clarity of harm reduction strategies. You will be provided with the original query posed by the health advice seeker (presented as ORIGINAL\_QUERY) and asked to the detect whether or not the harm reduction strategy suggested by the healthcare provider (presented as HARM\_REDUCTION\_STRATEGY) meets the criteria for being practical, contextually relevant, specific, and clear. We define each of the dimensions as follows:
         \\
         &Practicality: The proposed strategy should clearly identify at least one action the user can take. Further, it should be contextually feasible/practical, considering their personal circumstances, such as physical ability, financial resources, and time constraints.
         \\
         &Contextual relevance: The provided strategy should be relevant and should contribute to addressing the concern of the patient.
         \\
         &Specificity: The details and instructions provided in the harm reduction strategy should not be vague and should leave no to little room for risky interpretation.
         \\
         &Clarity: The strategy provided can be broken up into clear, manageable and explicit steps.
         \\
         \\
         &Make sure that you output the results strictly in the following format: \{'rationale\_to\_assess\_practicality': '<your rationale goes here>', 'practicality\_decision': '<0 or 1>', 'rationale\_to\_assess\_contextual\_relevance': '<your rationale goes here>', 'contextual\_relevance\_decision': '<0 or 1>', 'rationale\_to\_assess\_specificity': '<your rationale goes here>', 'specificity\_decision': '<0 or 1>', 'rationale\_to\_assess\_clarity': '<your rationale goes here>', 'clarity\_decision': '<0 or 1>'\} where 0 indicates that the strategy does not meet the criteria and 1 indicates that it does.
\\
&\\
&Consider the following examples:\\
&\\
&ORIGINAL\_QUERY: <example\_1\_query>\\
&HARM\_REDUCTION\_STRATEGY: <example\_1\_hrs>\\
&OUTPUT: \{'rationale\_to\_assess\_practicality': <example\_1\_practicality\_rationale>, 'practicality\_decision': <example\_1\_practicality\_decision>, \\&'rationale\_to\_contextual\_relevance':<example\_1\_contextual\_relevance>, 'contextual\_relevance\_decision': <example\_1\_contextual\_relevance\_decision>,\\ &'rationale\_to\_assess\_specificity':<example\_1\_specificity\_rationale>,'specificity\_decision':\\&<example\_1\_specificity\_decision>, \\&'rationale\_to\_assess\_clarity':<example\_1\_clarity\_rationale>,'clarity\_decision':<example\_1\_clarity\_decision>\}\\
&...\\
&...\\
&'rationale\_to\_assess\_clarity':<example\_3\_clarity\_rationale>,'clarity\_decision':<example\_3\_clarity\_decision>\}\\
         &\\
         \midrule
         \textbf{User Prompt}&ORIGINAL\_QUERY: <query>\\
         &HARM\_REDUCTION\_STRATEGY: <harm\_reduction\_strategy>\\
         \bottomrule
    \end{tabular}
    \caption{Prompt used for the decomposition of actionability in harm reduction strategies.}
    \label{tab:appendix_hrs_actionability_prompt}
\end{table*}

\end{document}